\pdfoutput=1

\documentclass{article}

\usepackage{times}
\usepackage{graphicx} 
\usepackage{subfigure} 

\usepackage{natbib}
\usepackage{booktabs}
\usepackage{algorithm}
\usepackage{algorithmic}

\usepackage{hyperref}

\usepackage[accepted]{icml2015a}
\usepackage{url}

\usepackage{epsfig}
\usepackage{epstopdf} 

\usepackage{comment}
\usepackage{amsmath,amssymb, amsthm}

\usepackage{multirow}
\usepackage{amsfonts}
\usepackage{layout}
\usepackage{amsmath}
\usepackage{amsthm}
\usepackage{amssymb}
\usepackage{url}
\usepackage{color}
\usepackage{graphicx}
\usepackage{algorithmic}
\usepackage{algorithm}
\usepackage{verbatim}
\usepackage{epsfig}
\usepackage{epstopdf} 
\usepackage{xcolor}

\PassOptionsToPackage{normalem}{ulem}
\usepackage{ulem}

\usepackage{cases}

\providecolor{added}{rgb}{0,0,1}
\providecolor{deleted}{rgb}{1,0,0}

\definecolor{orange}{RGB}{255,127,0}

\newcommand{\ve}[2]{\langle #1 ,  #2 \rangle}

\newcommand{\R}{\mathbb{R}}

\newcommand{\vc}[2]{#1^{[#2]}}

\newtheorem{assumption}{Assumption}

\theoremstyle{plain}

\theoremstyle{definition}

\newcommand*{\starnr}{\stepcounter{equation}\tag{\theequation}}

\usepackage{enumitem}

\icmltitlerunning{Distributed 
Inexact Damped Newton Method:
Data Partitioning and Work-Balancing}
\usepackage[colorinlistoftodos,bordercolor=orange,backgroundcolor=orange!20,linecolor=orange,textsize=scriptsize]{todonotes}

\graphicspath{{exp/log/}{./}{matlab/}}

\begin{document} 

\twocolumn[
\icmltitle{Distributed 
Inexact Damped Newton Method: Data
Partitioning and Load-Balancing}

\icmlauthor{Chenxin Ma}{chm514@lehigh.edu}
\icmladdress{Industrial and Systems Engineering, Lehigh University, USA}
\icmlauthor{Martin Tak\'a\v{c}}{Takac.MT@gmail.com}
\icmladdress{Industrial and Systems Engineering, Lehigh University, USA}

\icmlkeywords{SDCA, ERM, Adaptive Probabilities} 
]
 
\begin{abstract} 
In this paper we study 
inexact dumped Newton method implemented in a distributed environment.
We start with an original 
DiSCO algorithm 
[Communication-Efficient Distributed Optimization of
Self-Concordant Empirical Loss, Yuchen Zhang and
Lin Xiao, 2015].
We will show that this algorithm may not scale well and propose an algorithmic modifications
which will lead to less communications,  better load-balancing and more efficient computation.
We perform numerical experiments with an regularized empirical loss minimization 
 instance described by a  273GB dataset.

\end{abstract}

\section{Introduction}

As the size of the datasets becomes larger and larger, distributed optimization methods for machine learning have become increasingly important \cite{bertsekas1989parallel,dekel2012optimal,shamir2014distributed}. Existing mehods often require a large amount of communication between computing nodes \cite{yang2013trading,jaggi2014communication,ma2015adding,yang2013analysis}, which is typically several magnitudes slower than reading data from their own memory \cite{marecek2014distributed}. Thus, distributed machine learning suffers from the communication bottleneck on real world applications.

In this paper we focus on the regularized empirical risk minimization problem. Suppose we have $n$   data samples $\{x_i, y_i\}_{i=1}^n$,
where each $x_i \in \R^d$ (i.e. we have $d$ features), $y_i \in \R$.
We will denote by $X$ the data matrix, i.e. $X := [x_1,...,x_n]\in\R^{d\times n}$. The optimization problem is to minimize  the regularized empirical loss  
\begin{equation}\label{eq:pro1}
  f(w) := \frac{1}{n} \sum_{i=1}^n \phi_i(w,x_i) + \frac{\lambda}{2} \|w\|_2^2,
  \tag{P}
\end{equation}
where the 
first part is the \emph{data fitting term}, 
 $\phi: \R^d\times \R^d \rightarrow \R$ is 
 L-smooth  
 loss function which typically depends on $y_i$.
 \footnote{Function $\phi$ is $L$-smooth, if 
 $\nabla \phi(\cdot)$ is $L$-Lipschitz continuous.}
 The second part of objective function \eqref{eq:pro1} is $\ell_2$ regularizer ($\lambda>0$)
 which helps to prevent over-fitting of the data.

There has been an enormous interest in large-scale machine learning problems and many parallel \cite{bradley2011parallel,recht2011hogwild}
or distributed algorithms have been proposed 
\cite{agarwal2011distributed,takavc2015distributed,
richtarik2013distributed,shamir2013communication,
lee2015distributed}. 

From algorithmic point of view some researches try to minimize  \eqref{eq:pro1} 
directly including SGD \cite{shalev2011pegasos},
SVRG and S2GD \cite{johnson2013accelerating,
nitanda2014stochastic,
konevcny2014ms2gd} and SAG/SAGA
\cite{schmidt2013minimizing,defazio2014saga,
roux2012stochastic}.
On the other side, one very popular  approach is to   solve its dual formulation  \cite{hsieh2008dual}
which has been successfully done in multicore or distributed settings 
\cite{
takavc2013mini,
jaggi2014communication,
ma2015adding,
takavc2015distributed,
qu2015quartz,
csiba2015stochastic,
zhang2015communication}.
The dual problem of \eqref{eq:pro1} has following form:
\begin{equation} 
\tag{D} \label{Prob:dual}
\max_{\alpha \in \R^n} D(\alpha) :=
-\frac1n \sum_{i=1}^n \phi^*_i(-\alpha_i) -\frac{\lambda}{2} \|\frac1{\lambda n} X  \alpha\|^2,
\end{equation}
$\phi_i^*$ is a convex conjugate function of $\phi_i$.

\paragraph{The Challenge In Distributed Computing.}
We can identify few challenges when we deal with high-performance distributed environment.  
\begin{enumerate}[topsep=0pt,itemsep=-1ex,partopsep=1ex,parsep=1ex]
\item {\bf Load-Balancing.} Assume that we have $m$ computational nodes available use. In order to have an algorithm, which is scalable, the algorithm should make each node equally busy. Amdahl's law \cite{rodgers1985improvements}
implies that if the parallel/distributed algorithm runs e.g. 75\% of the time only on one of the nodes (usually the master node), then
the possible   speed-up of the algorithm is bounded by 
$
\frac{1}{0.75+0.25/m}$ and is shown on Figure \ref{fig:speed-up} and is asymptotically bounded by $4/3\approx 1.333$.

Hence, any algorithm which is targeted for a very large scale problems has to be design in such a way, that the sequential portion of the algorithm is negligible.
\begin{figure}[h]
\centering
\includegraphics[scale=0.3]{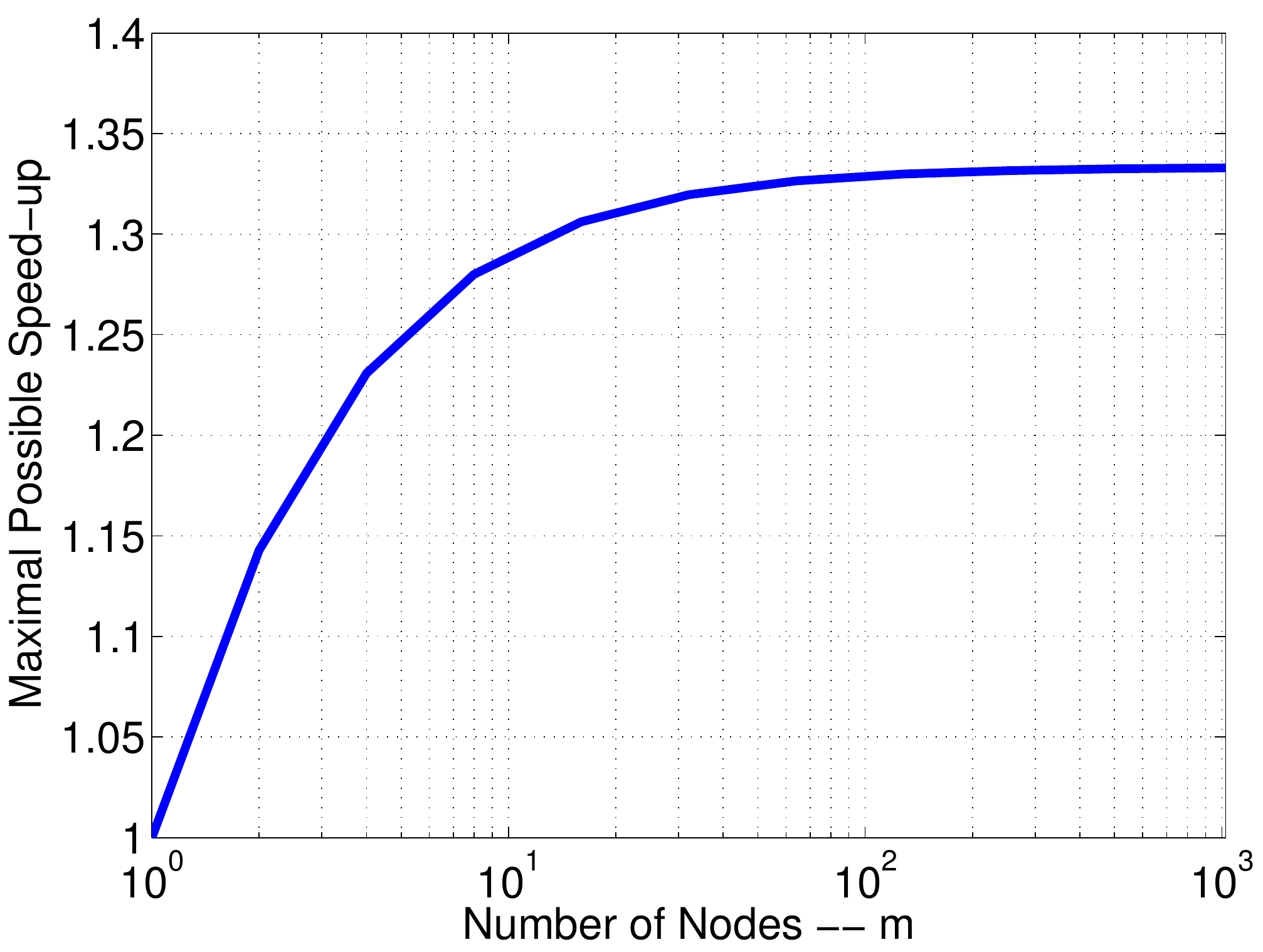}
\caption{Maximal possible speed-up of an Algorithm which runs 75\% in sequential mode.}
\label{fig:speed-up}
\end{figure}
\item {\bf Communication efficiency.}
As it was stressed in the introduction, in a distributed setting the communication between nodes should be avoided or minimized (if possible).
Another challenge is to balance the time the nodes are going some computation and the time they spent in the communication (usually in MPI calls).
 
\end{enumerate}
 
In this paper we modify the design  of promising DiSCO algorithm \cite{zhang2015communication}.
We completely redesign the algorithm (partitioning of the data, preconditioning, communication patterns)
to get a new algorithm which 
\begin{enumerate}[topsep=0pt,itemsep=-1ex,partopsep=1ex,parsep=1ex]
\item has very good scaling (the serial computation is almost negligible),
\item balances work-load equally between nodes (all the nodes are working all the time, no real job for master node),
\item has much smaller amount of data set over the network  than in the original DiSCO algorithm.
\end{enumerate}

\subsection{Related Work}

As stressed in the previous section, one of the  main bottlenecks in distributed computing is  communication.
This challenge  was
handled by many researchers differently. 
In the ideal case, 
one would like to never communicate 
or maybe communicate only once at the end to form a result. However, such a procedure which could communicate only max once and would be able to give arbitrary good solutions (when no node can have access to all the data) is more fantasy than reality.

Hence, to somehow "synchronize" work on different computing nodes, researchers use various standard technique from optimization.
Few of them based their algorithms on  ADMM type methods \cite{boyd2011distributed,deng2012global},
another used block-coordinate type algorithms
\cite{lee2015distributed,yang2013trading,jaggi2014communication,ma2015adding}, where they solved on each node some local sub-problems
which together formed an upper-bound on the optimization problem.
The balancing of computation and communication  
was achieved by varying the accuracy of the solutions of the local sub-problems, which turns out to be more efficient than some earlier approaches \cite{takavc2013mini,takavc2015distributed}).

Let us now give an overview of existing approaches to solve problem \eqref{eq:pro1} in distributed setting:
\begin{enumerate}[topsep=0pt,itemsep=-1ex,partopsep=1ex,parsep=1ex]
\item 

{\bf SGD-based Algorithms.}
For the empirical loss minimization problems of interest here, stochastic subgradient descent (SGD) based methods are well-established.
Several distributed variants of SGD have been proposed, many of which build on the idea of a parameter server \cite{Niu:2011wo,Liu:2014wj,Duchi:2013te}. 
The downside of this approach, even when carefully implemented, is that the amount of required communication is equal to the amount of data read locally (e.g., mini-batch SGD with a batch size of 1 per worker). These variants are in practice not competitive with the Newton-type methods considered here, which allow more local updates per communication round.

\item 

{\bf ADMM.}
An alternative approach to distributed optimization is to use the alternating direction method of multipliers (ADMM), as used for distributed SVM training in, e.g., \cite{forero2010consensus}. This uses a penalty parameter balancing between the equality constraint  and the optimization objective \cite{boyd2011distributed}. However, the known convergence rates for ADMM are weaker than the more problem-tailored methods mentioned previously, and the choice of the  penalty parameter is often unclear.

\item 

{\bf DANE Algorithm.} 

The DANE (Distributed Approximate Newton) algorithm \cite{shamir2013communication} is applicable to solve problems \eqref{eq:pro1} with any smooth loss functions, which genreally requires two rounds of communication in each iteration. In $k$-th iteration, the first round of communication is used to compute the gradient $f'(w_k)$ by a ReduceAll operation on a $\R^d$ vector. Then each machine solves the local problem 
\begin{align}\label{eq:danesubp}
  w_i = \arg\min_w\{f_i(w) - &(\nabla f'_i(w_k) - \eta \nabla f(w_k))^T w \notag\\
  & + \frac{\mu}{2}\|w-w_k\|_2^2\},
\end{align}
and takes the second round of communication to compute $w_{k+1}$ by another ReduceAll operation. The parameter $\mu\geq0$ plays the  role of damping. The iteration complexity of DANE is $\mathcal O(\frac1n (\frac L\lambda)^2\log \frac1\epsilon )$, if the loss function in \eqref{eq:pro1} is quadratic loss.  However, current analysis does not guarantee that DANE has the same convergence rate on non-quadratic functions.

\item 

{\bf CoCoA+ Algorithm.} 
The CoCoA+ (Communication Efficient Primal-Dual Coordinate Ascent Framework) \cite{jaggi2014communication,ma2015distributed,ma2015distributed} allows each machine to solve the subproblem which can be regarded as a variant of dual problem of \eqref{eq:pro1}. It allows additive combination of local updates to the global parameters at each iteration, and requires only one round of communication per iteration. Also, the trade-off between communication and local computation can be controled freely based on the problem and the system hardware. CoCoA+ uses only first order information, and the primal-dual convergence rate has been proven for both smooth and general convex (L-Lipschitz) loss functions \cite{ma2015adding}. The iteration complexity for smooth loss functions is $\mathcal O(\frac{L}{\lambda n} \log \frac1\epsilon)$.

\item {\bf DiSCO Algorithm.}
The DiSCO algorithm 
is a Newton type method, where in each iteration the step is  solved inexactly using Preconditioned Conjugate Gradient (PCG) method.
It, however, requires the step to be computed to some good accuracy. 
Number of iterations of PCG depends on many factors, 
including the data partitioning, the quality of pre-conditioner 
and of course, the requested desired accuracy.
Number of PCG steps also changes from iteration to iteration as the local geometry of the problem changes.
Compared with other methods discussed above, few parameters need to be tuned for optimizing the performance in DiSCO. 

\end{enumerate}

\subsection{Contributions}
In this section we summarize the main contributions of this paper (not in order of significance).

\begin{enumerate}
\item {\bf Preconditioning is solved in closed form and efficiently.}
The PCG methods need to solve the 
the preconditioned system of linear equations.
In general, solving the linear system exactly is very expensive due to the dimension of the problem, especially when $d$ is large even infeasible.
Therefore, in 
DiSCO algorithm
\cite{zhang2015communication}
authors suggested to use an iterative method to solve it. They suggested to use SAG/SAGA algorithm which has a linear rate of convergence. However, the SAG/SAGA algorithm is run only on master node, while all other workers are being idle, and unfortunately, the time to solve the preconditioned system is not at all negligible. In our experiments, we observed that for same dataset the percentage of time spent in solving PCG was more then 50\% which implies  poor scaling of the DiSCO algorithm.

To overcome that issue, we propose a  new   preconditioning matrix $P$, which can be viewed as an approximated or stochastic Hessian. 
By exploring the structure of the new preconditioning matrix $P$, the linear system can be solved much more efficient (actually, exactly) by Woodbury Formula.
Because, the matrix $P$ is constructed only based on  $\tau\ll n$ samples, the time needed to  solve the preconditioning system is negligible.  
By applying this approach, we proposed a variant of DiSCO algorithm called DiSCO-S. Our practical experiments in Section \ref{sec:numExperiments} not only confirms that this preconditioning is superior to the preconditioning suggested in original DiSCO algorithm,
but also demonstrate that a very small $\tau$ would give a good performance.

\item {\bf Data Partitioned by Features.}
In our setting we assume that the dataset is large enough that it cannot be stored entirely on any single node and hence the dataset has to be partitioned.

Both DiSCO and DiSCO-S algorithms are based on partitioning dataset by samples. By considering another way of making partitions, i.e., partitioning by features, we proposed a new DiSCO-F algorithm. In this new setting, the number of communications is reduced by half compared with the original DiSCO. Moreover, the computation in each machine is more balanced, such that the computing resources can be better utilized. Compared with making partitions by samples, we do not need to pick a machine as the master node, which will do more computation than others. In the DiSCO-F algorithm, all machines will do exactly the same work and the computation will be distributed more properly,
hence it can possible obtain almost linear speed-up. 


\end{enumerate}

\section{Assumptions}

We assume that the loss function $\phi_i$ is convex and  self-concordant \cite{zhang2015communication}:
\begin{assumption}\label{ass:selfc}
For all $ i \in [n]:=\{1,2,\dots,n\}$ the convex function $\phi$ is self-concordant with parameter $M$ i.e. the following inequality holds:
\begin{equation}
  |u^T (f'''(w)[u])u| \leq M(u^T f''(w) u) ^{\frac{3}{2}}
\end{equation}
for any $u\in\R^d$ and $w\in dom(f)$, where $f'''(w)[u]:= \lim_{t\rightarrow 0} \frac{1}{t} \large(f''(w+tu)- f''(w)\large)$.
\end{assumption}

Table \ref{tab:losses}
lists some examples of loss functions which satisfy the Assumption \ref{ass:selfc}
with corresponding constant $M$.
\begin{table}[H]
 \caption{Loss functions satisfying Assumption \ref{ass:selfc} and the parameter $M$.}
  \label{tab:losses}%
  \centering
    \begin{tabular}{c|c|c}
      & $\phi_i(w,x_i)$  & $M$ \\
    \hline
   quadratic loss  & $(y_i - w^T x_i)^2 $  & 0\\
   \hline
  squared hinge loss  & $(\max\{0, y_i - w^T x_i\})^2$  & 0\\
   \hline
    logistic loss  & $ \log(1+\exp(-y_iw^T x_i)) $  & 1\\
    \end{tabular}%
 
\end{table}%

 Also, we assume that the function $f$ is both $L$-smooth and $\lambda$-strongly convex.
\begin{assumption}  \label{ass:fsecondassume}
The function $f:\R^d\rightarrow \R$ is trice continuously differentiable, and there exist constants $L\geq \lambda >0$ such that 
\begin{equation}
\lambda I\preceq f''(w)\preceq LI, \quad\forall w\in\R^d,
\end{equation}
\end{assumption}
\textit{where $f''(w)$ denotes the Hessian of $f$ at $w$, and $I$ is the $d\times d$ identity matrix. }

\begin{table*}[htbp]
  \centering
  \caption{Communication efficiency of several distributed algorithms when the regularization parameter $\lambda \sim 1/\sqrt{n}$ (common seen in machine learning), we can simplify and compare the complexity for the three algorithms. CoCoA+ use more round of communication to reach $\epsilon$ accuracy since it is a first order method. DANE and DiSCO are Newton-type methods, which tend to use less communications.}
    \begin{tabular}{rcc}
    \toprule
    \multicolumn{1}{c}{\multirow{2}[0]{*}{Algorithm}} & \multicolumn{2}{c}{Number of Communication } \\
    \multicolumn{1}{c}{} & Quadratic Loss & Logistic Loss \\
     \midrule
   DANE  &   $m\log(1/\epsilon)$    & $(mn)^{1/2}\log(1/\epsilon)$  \\
  CoCoA+   &    $n\log(1/\epsilon)$    & $n\log(1/\epsilon)$  \\
  DiSCO   &    $m^{1/4} \log(1/\epsilon)$   & $m^{3/4}d^{1/4}+ m^{1/4}d^{1/4}\log(1/\epsilon)$ \\
    \bottomrule
    \end{tabular}%
  \label{tab:addlabel}%
\end{table*}%

 \setlength{\textfloatsep}{7pt}

\section{Algorithm}

We assume that we have  $m$ machines (computing nodes) available
which can communicate between each other over the network.
We assume that the space needed to store the data matrix $X$ exceeds the memory of every single node. Thus we have to split the data (matrix $X$) over the $m$ nodes. The natural question is: \emph{How to split the data into $m$ parts?}
There are many possible ways, but two obvious ones:
\begin{enumerate}
[topsep=0pt,itemsep=-1ex,partopsep=1ex,parsep=1ex]
\item split the data matrix $X$ by rows (i.e.  create $m$ blocks by rows);
Because rows of $X$ corresponds to features, we will denote the algorithm which is using this type of partitioning as \emph{DiSCO-F};

\item split the data matrix $X$ by columns;
Let us note that columns of $X$ corresponds to samples  we will denote the algorithm which is using this type of partitioning as \emph{DiSCO-S};
\end{enumerate}
 \begin{algorithm}[H]
\caption{High-level DiSCO algorithm}
\label{Disco}
\begin{algorithmic}[1]
\STATE\textbf{Input: } parameters $\rho, \mu \geq 0$, number of iterations $K$
\STATE Initializing $w_0$.
\FOR  {k = 0,1,2,...,K}
\STATE Option 1: Given $w_k$, run DiSCO-S PCG Algorithm \ref{A1}, get $v_k$ and $\delta_k$
\STATE Option 2: Given $w_k$, run DiSCO-F PCG Algorithm \ref{A2}, get $v_k$ and $\delta_k$
\STATE Update $w_{k+1} = w_k - \frac{1}{1+ \delta_k} v_k$
\ENDFOR
\STATE \textbf{Output: $w_{K+1}$}
\end{algorithmic}
\end{algorithm}
Notice that the DiSCO-S is exactly the same as DiSCO proposed and analyzed 
 in \cite{zhang2015communication}. 
 In each iteration of Algorithm \ref{Disco}, wee need to compute an inexact Newton step $v_k$ such that $\|f''(w_k)v_k - \nabla f'(w_k)\|_2 \leq \epsilon_k$, which is an approximate solution to the Newton system $f''(w_k)v_k = \nabla f (w_k)$. The discussion about how to choose  $\epsilon_k$ and $K$ and a convergence guarantees  for  Algorithm \ref{Disco} can be found   in \cite{zhang2015communication}. And the main convergence result still applies here: 
If Algorithm \ref{A1} or \ref{A2} is run starting with $w^0$
then after
$$ T\sim\mathcal O \Big((f(w^0) - f(w^*) + \log(1/\epsilon)) \sqrt{1+2\mu/\lambda} \Big)$$
communication rounds (iterations)
the algorithm will produce a solution  $\hat w$ satisfying $f(\hat w) - f(w^*) <\epsilon$.

The main goal of this work is to analyze the algorithmic modifications to DiSCO-S when the partitioning type is changed. It will turn out that partitioning on features (DiSCO-F) can lead to an algorithm which uses less communications (depending on the relations between $d$ and $n$) (see Section \ref{sec:numExperiments}).

 \begin{algorithm*}[t!]
\caption{Distributed DiSCO-S: PCG algorithm -- data  partitioned by  samples}
\label{A1}
\begin{algorithmic}[1]
\STATE\textbf{Input: } $w_k\in \R^{d}$, and $\mu \geq 0$.\hfill communication (Broadcast $w_k\in\R^d$ and reduceAll $\nabla f_i(w_k)\in\R^d$ )
\STATE\textbf{Initialization: } Let $P$ be computed as \eqref{eq:precon}. $v_0 = 0$, $s_0 = P^{-1} r_0$, $r_0 =\nabla f(w_k)$, $u_0 = s_0$.
\FOR {$t= 0,1 ,2,...$}
\STATE Compute $Hu_t$       \hfill communication (Broadcast $u_t\in\R^d$ and reduceAll $f{''}_i(w_k)u_t\in\R^d$ )
\STATE {Compute $\alpha_t = \frac{ \ve{r_t}{s_t} }{\ve{u_t}{Hu_t} }$}  
\STATE Update $v_{(t+1)} = v_t+ \alpha_t u_t, Hv_{(t+1)} = Hv_t + \alpha_t Hu_t, r_{t+1} = r_t - \alpha_t H u_t$.
\STATE Update $ P s_{(t+1)} =  r_{(t+1)}$.
\STATE {Compute $\beta_t = \frac{  \ve{r_{(t+1)} }{ s_{(t+1)} }}{ \ve{r_t}{s_t} }$}
\STATE Update  $u_{(t+1)} = s_{(t+1)} + \beta_t u_t$.
\STATE\textbf{until: $\|r_{(r+1)}\|_2 \leq \epsilon_k$}
\ENDFOR
\STATE \textbf{Return: $v_k =v_{t+1}$, $\delta_k = \sqrt{v_{(t+1)}^THv_{t} + \alpha_t v_{(t+1)}^T Hu_t}$}
\end{algorithmic}
\end{algorithm*}

\paragraph{DiSCO-S Algorithm.} 
If the dataset is partitioned by samples, such that $j$--th node will only store $X_j =[x_{j,1},...,x_{j,n_j}] \in\R^{d\times n_j}$, which is a part of $X$, then each machine can evaluate a local empirical loss function
\begin{equation}
   f_j(w):= \frac{1}{n_j} \sum_{i=1}^{n_j} \phi(w,x_{j,i}) + \frac{\lambda}{2} \|w\|_2^2.
\end{equation} 
Because $\{X_j\}$ is a partition of $X$ we have 
  $\sum_{j=1}^m n_j = n$, our goal now becomes to minimize the function $f(w) = \frac{1}{m} \sum_{h=1}^{m} f_j(w)$. Let $H$ denote the Hessian $f''(w_k)$.
For simplicity in this section we present it only for  square loss  (and hence in this case $f''(w_k)$ is constant -- independent on $w_k$), however, 
it naturally extends to any smooth loss.

In Algorithm \ref{A1}, each machine will use its local data to compute the local gradient and local Hessian and then aggregate them together. We also have to choose one machine as the master, which computes all the vector operations of PCG loops (Preconditioned Conjugate Gradient), i.e., step 5-9 in Algorithm \ref{A1}. 

The preconditioning matrix for PCG is defined only on master node and consists of the local Hessian approximated by a subset of data available on master node with size $\tau$, i.e.
\begin{equation}\label{eq:precon}
  P =   \frac{1}{\tau} \sum_{j=1}^\tau \phi^{''} (w, x_{1,j}) + \mu I,
\end{equation}
where $\mu$ is a small regularization parameter. Algorithm \ref{A1} presents the distributed PCG mathod for solving the  linear system
\begin{equation}\label{eq:llllllll}
   H v_k =  \nabla f(w_k).
\end{equation}

\begin{algorithm*}[t!]
\caption{Distributed DiSCO-F: PCG algorithm -- data  partitioned by features}
\label{A2}
\begin{algorithmic}[1]
\STATE\textbf{Input: } $\vc{w_k}{j}\in \R^{d_j}$ for $j = 1,2,...,m$, and $\mu \geq 0$. 
\STATE\textbf{Initialization: } Let $P$ be computed as \eqref{eq:precon}. $\vc{v_0}{j} = 0$, $\vc{s_0}{j} = {(P^{-1})}^{[j]} \vc{r_0}{j}$, $\vc{r_0}{j} = f'(\vc{w_k}{j})$, $\vc{u_0}{j} = \vc{s_0}{j}$.
\WHILE {$\|r_{r+1}\|_2 \leq \epsilon_k$}
\STATE Compute $\vc{(Hu_t)}{j}$. \hfill communication (ReduceAll an $\R^{n}$ vector)
\STATE {Compute $\alpha_t = \frac{\sum_{j=1}^{m}  \ve{\vc{r_t}{j}}{\vc{s_t}{j}} }{\sum_{j=1}^{m}  \ve{\vc{u_t}{j}}{\vc{(Hu_t)}{j}} }$}. \hfill communication (ReduceAll a number)
\STATE Update $\vc{v_{t+1}}{j} = \vc{v_t}{j} + \alpha_t \vc{u_t}{j}, \vc{(Hv_{t+1})}{j} = \vc{(Hv_t)}{j} + \alpha_t \vc{(Hu_t)}{j}, \vc{r_{t+1}}{j} = \vc{r_t}{j} - \alpha_t \vc{(Hu_t)}{j}$.
\STATE Update $P^{[j]}\vc{s_{t+1}}{j} =  \vc{r_{t+1}}{j}$.
\STATE {Compute $\beta_t = \frac{\sum_{j=1}^{m}  \ve{\vc{r_{t+1}}{j}}{\vc{s_{t+1}}{j}} }{\sum_{j=1}^{m}  \ve{\vc{r_t}{j}}{\vc{s_t}{j}} }$}. \hfill communication (ReduceAll a number)
\STATE Update  $\vc{u_{t+1}}{j} = \vc{s_{t+1}}{j} + \beta_t \vc{u_t}{j}$.
\ENDWHILE
\STATE Compute $\vc{\delta_k}{j}= \sqrt{{\vc{v_{t+1}}{j}}^T(Hv_t)^{[j]} + \alpha_t{\vc{v_{t+1}}{j}}^T (Hu_t)^{[j]}}$.
\STATE \textbf{Integration:  $v_{k}=[\vc{v_{t+1}}{1},...,\vc{v_{t+1}}{m}]$, $\delta_{k}=[\vc{\delta_{t+1}}{1},...,\vc{\delta_{t+1}}{m}]$} \hfill communication (Reduce an $\R^{d_j}$ vector)
\STATE \textbf{Return: $v_{k}$, $\delta_{k}$}
\end{algorithmic}
\end{algorithm*}

Notice that in Algorithm \ref{A1}, there is another linear system 
\begin{equation}\label{precon2}
  s=P^{-1}r
\end{equation}
to be solved, which has the same dimension as \eqref{eq:llllllll}. However, becasue we only apply a subset of data to compute the preconditioning matrix $P$, \eqref{precon2} can be solved by Woodbury formula \cite{press2007numerical}, which will be described detail in Section \ref{sec:wood}.

\paragraph{DiSCO-F Algorithm.} 
If the dataset is partitioned by features, then $j$-th machine will store $X_j = [\vc{a_1}{j},...,\vc{a_n}{j}]\in\R^{d_j\times n}$, which contains all the samples, but only with a subset of features. Also, each machine will only store $\vc{w_k}{j}\in\R^{d_j}$ and thus only be responsible for the computation and updates of $\R^{d_j}$ vectors. By doing so, we only need one ReduceAll on a vector of length $n$, in addition to two ReduceAll on scalars number.

\begin{figure}[t]
\centering
\includegraphics[scale=.27]{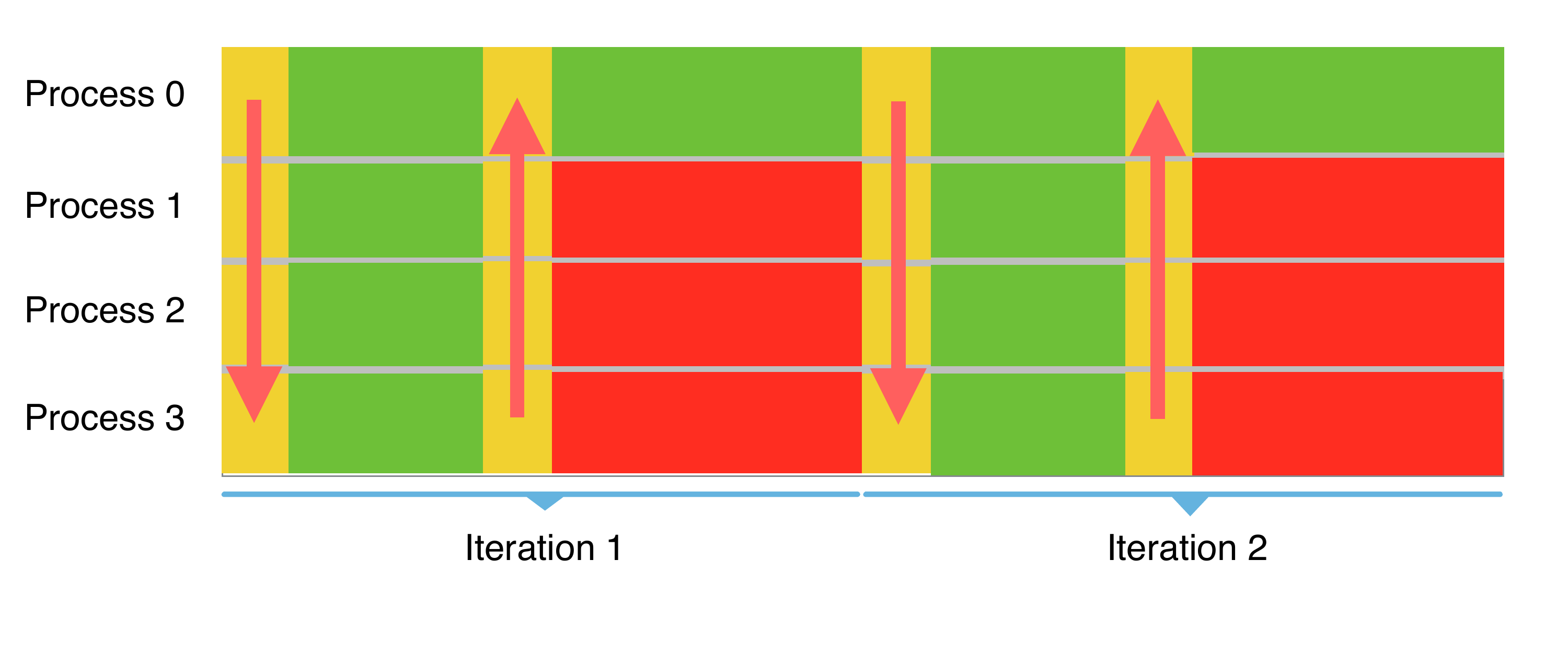}
\includegraphics[scale=.27]{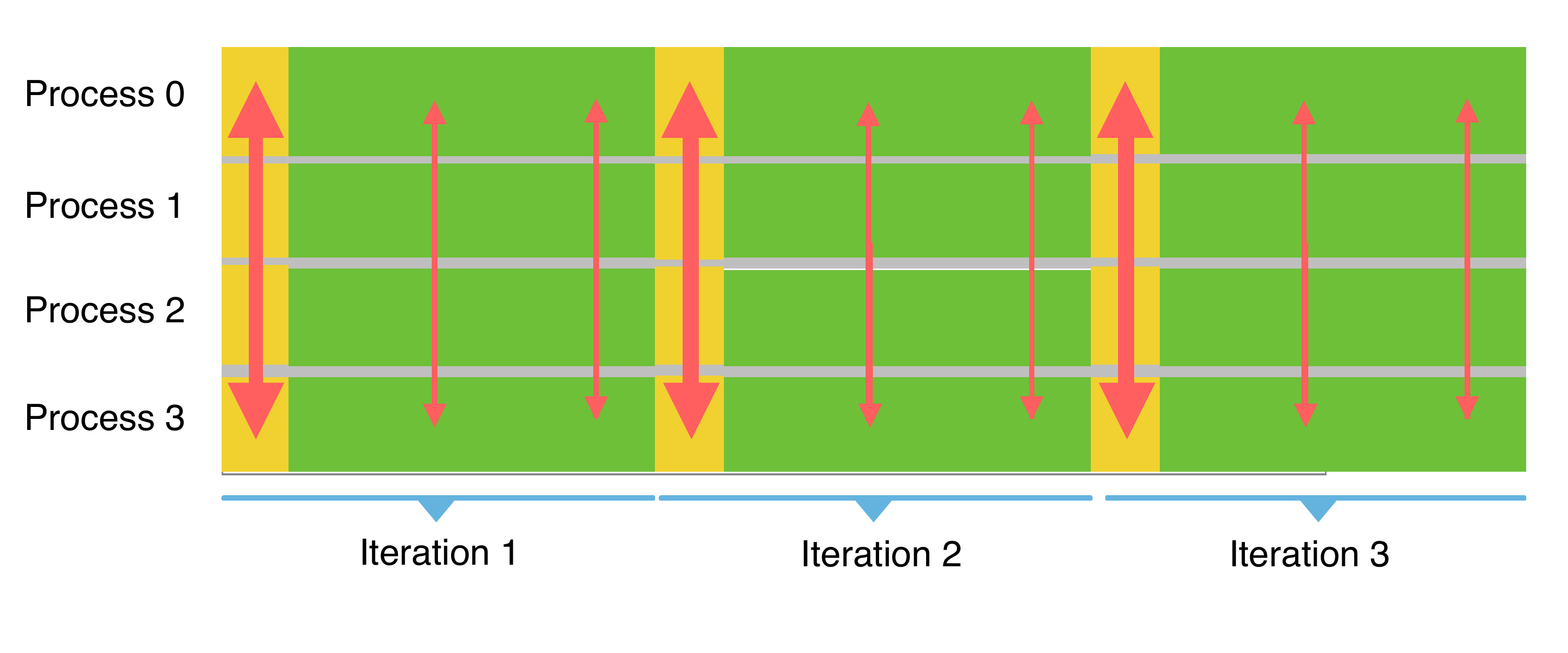}
\caption{Flow diagrams of few iterations of DiSCO-S (top) and DiSCO-F (bottom). DiSCO-F uses less time for one iteration, due to the more efficient and balanced computation. Green boxes represent the processes are busy, while red boxes represent idle nodes. Yellow boxes show the status of communicating between all processes. Double arrows stand for ReduceAll operations. The thin red arrows represent a communication of few scalars only.
} 
\label{fig:compareRcv}

\end{figure}

\paragraph{Comparison of Communication and Computational Cost.}  
In Table \ref{tab:commuFS} we compare the communication cost for the two approaches DiSCO-S/DiSCO-F. As it is obvious from the table, DiSCO-F 
requires only one reduceAll of a vector of length $n$, whereas the DiSCO-S needs one reduceAll of a vector of length $d$ and one broadcast of vector of size $d$. So roughly speaking, when $n < d$ then DiSCO-F will need less communication.
However, very interestingly, the advantage of DiSCO-F is the fact that it uses CPU on every node more effectively. It also requires less total amount of work to be performed on each node, leading to more balanced and efficient utilization of nodes (see Figure \ref{fig:compareRcv} for illustration how .
   DiSCO-F utilizes resources more efficiently
and Table \ref{tab:tsssssss} for the size of communication required in each PCG step).

\begin{table}[htbp]
 \caption{Comparison of computation between different algorithms. For DiSCO-S, there exists large differences of computation between the master node and other node. But for DiSCO-F, each node will do exactly the same computation as each other.}
  \label{tab:commuFS}%
  \centering
    \begin{tabular}{c|c|c|c}
    & Operation& DiSCO-S & DiSCO-F \\
    \hline
    \multirow{4}[0]{*}{master} & $y = Mx$ & $ 1 (\R^{d\times d}\cdot \R^d)$ &$1 (\R^{d_1\times d_1}\cdot \R^{d_1})$\\
           &$Mx=y$ & 1 ($\R^d$)  & 1 $(\R^{d_1})$ \\
           &$x+y$ & 4 ($\R^d$)  & 4 ($\R^{d_1}$)\\
          &$x^Ty$& 4 ($\R^{d}$) & 4 ($\R^{d_1}$)\\
        \hline
          {\multirow{4}[0]{*}{nodes}} & $y = Mx$  & 1 $ (\R^{d\times d}\cdot \R^d)$& $1 (\R^{d_1\times d_i}\cdot \R^{d_i})$ \\
            &$Mx=y$ &  0 & 1 $(\R^{d_i})$ \\
         &$x+y$     &0  &  4 $(\R^{d_i})$\\
          &$x^Ty$& 0 & 4 $(\R^{d_i})$ \\
    \end{tabular}%
\end{table}%

\begin{table}[H]
 \caption{Comparison of communication between different algorithms. }
  \label{tab:tsssssss}%
  \centering
    \begin{tabular}{c|c|c|c}
  DiSCO-S  & DiSCO-F  & DANE & CoCoA+  \\
     \hline
      $2\times\R^d$  & $1\times\R^n, 2\times\R$ &  $2\times\R^d$&  $1\times\R^d$\\
    \end{tabular}%
 
\end{table}%


\section{Woodbury Formula for solving $Ps = r$}\label{sec:wood}

In each iteration of Algorithms \ref{A1} and \ref{A2}, we need to solve a linear system in the form of $Ps= r$, where $P\in\R^{d\times d}$ in Algorithm \ref{A1} and $P\in\R^{d_i \times d_i}$ for $i=1,2,...,m$ in Algorithm \ref{A2}, which is usually very expensive. To solve it more efficiently, we can apply Woodbury Formula \cite{press2007numerical}. 

Notice that if we use $P$ defined in \eqref{eq:precon}, $P$ can be considered as $\tau$ rank-1 updates on a diagnal matrix. For example, if $\phi(\cdot)$ is Quadtratic Loss, then 
\begin{equation}
  P = D + \frac{1}{\tau}\sum_{i=1}^\tau x_i x_i^T.
\end{equation}
If $\phi(\cdot)$ is Logistic Loss, then
\begin{equation}
  P = D + \frac{1}{\tau}\sum_{i=1}^\tau \frac{\exp(-w_k^T x_i)}{(\exp(-w_k^T x_i)+1)^2}x_i x_i^T.
\end{equation}
In both cases, $D$ is the diagnal matrix with $D_{ii} = \lambda + \mu$ for $i = 1,...,m$. Then we can follow the procedure to get the solution $s$.

\begin{algorithm}[H]
\caption{Woodbury Formula to solve $Ps = r$}
\label{A2222}
\begin{algorithmic}[1]
\STATE Compute $z_i = \frac{1}{\lambda+\mu} x_i$ for $i = 1,...,\tau$
\STATE Let $Z = [z_1,...,z_\tau]$, $X = [x_1,...,x_\tau] $
\STATE Compute $y = \frac{1}{\lambda+\mu} r$ 
\STATE Solve the linear system $(I+ X^T Z) v= X^T y$
\STATE \textbf{Return}: $s = y - X v$
\end{algorithmic}
\end{algorithm}

Notes that $v\in\R^\tau$ and $\tau\ll d$ (in our experiments, $\tau = 100$ usually works very well), step 4 can be done efficiently by any solver. In Section \ref{sec:expP}, we compare the effect of setting different values for $\tau$.

\section{Numerical Experiments}  
  \label{sec:numExperiments}

We present experiments on several large real-world datasets distributed across multiple machines, running on an Amazon EC2 cluster with 4 instances. We show that DiSCO-F with a small $\tau$ converges to the optimal solution faster in terms of total rounds of communications compared to original DiSCO, DANE and CoCoA+ in most cases. Also, for the dataset with $d>n$, DiSCO-F will also dominate others in elapsed time. In Section \ref{sec:expP}, we compare the affects of preconditioning matrices with different values for $\tau$, and investigate that a small $\tau$ (around 100) would result in a impressive performance. In Section \ref{sec:expHess}, we show that extra speed-up can be gained in some cases, by trying to shrink the number of samples that are used to compute Hessian.
 
\begin{figure*}[t]
\centering
\includegraphics[scale=.15]{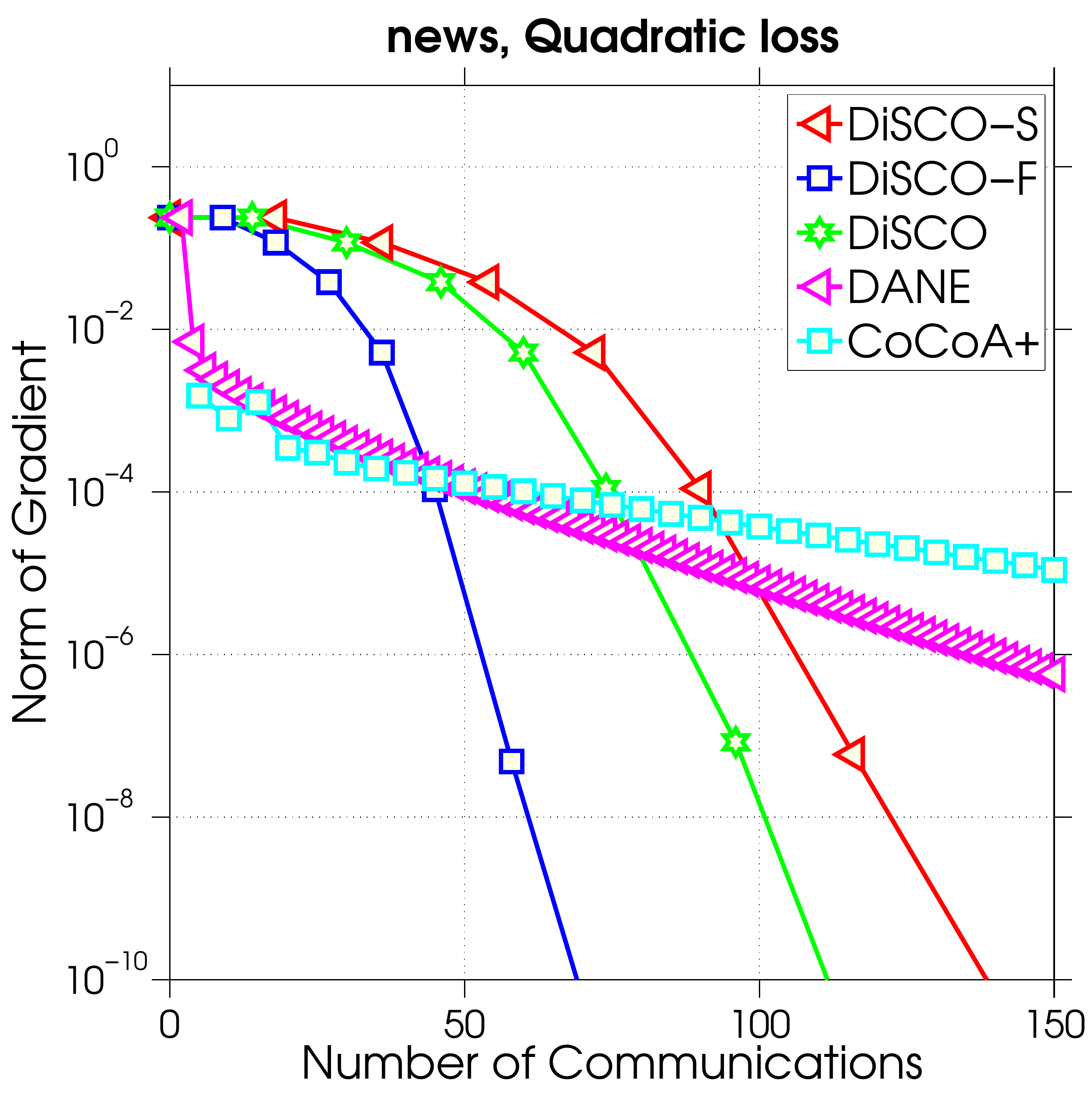}
\includegraphics[scale=.15]{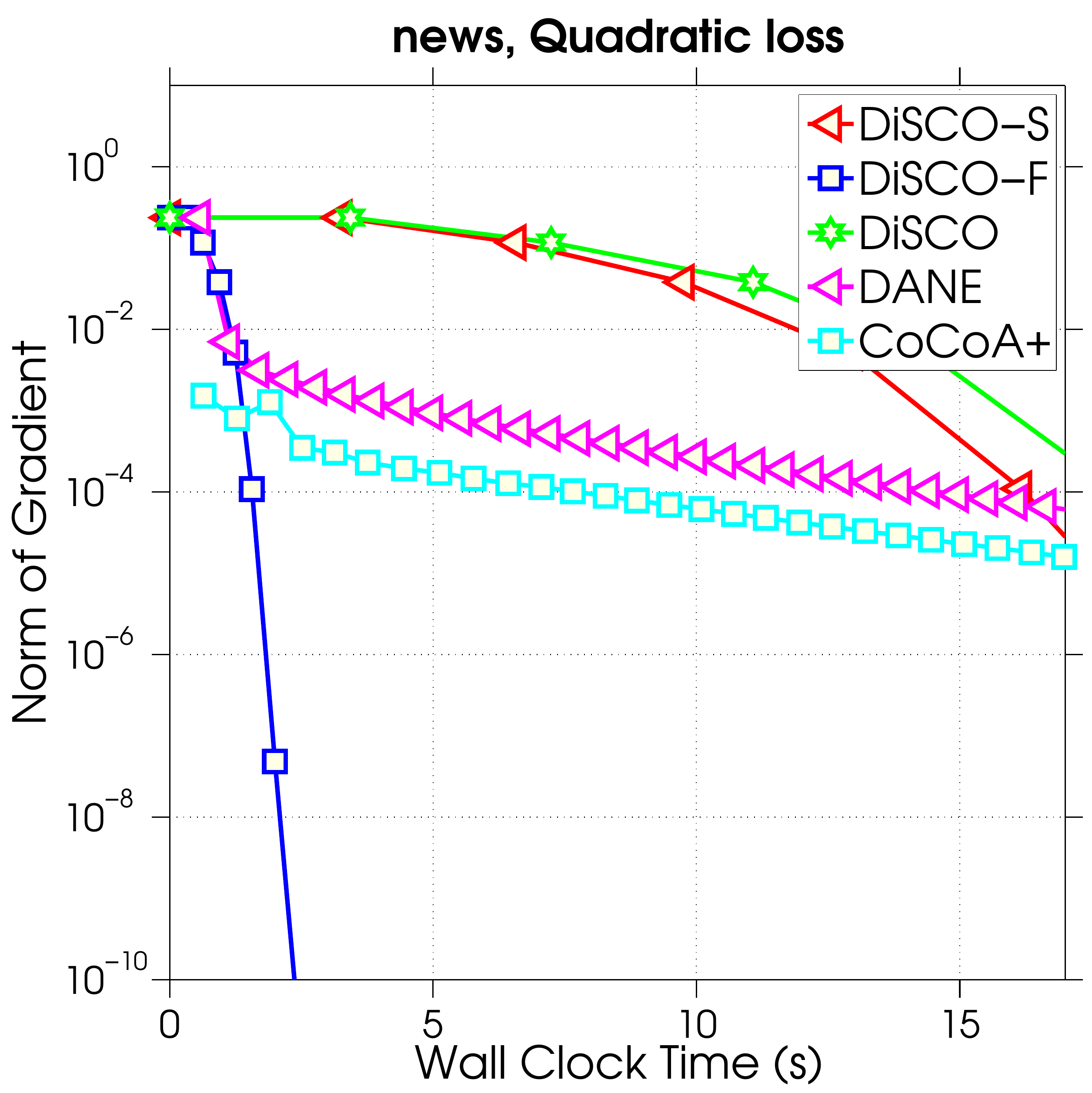}
\includegraphics[scale=.15]{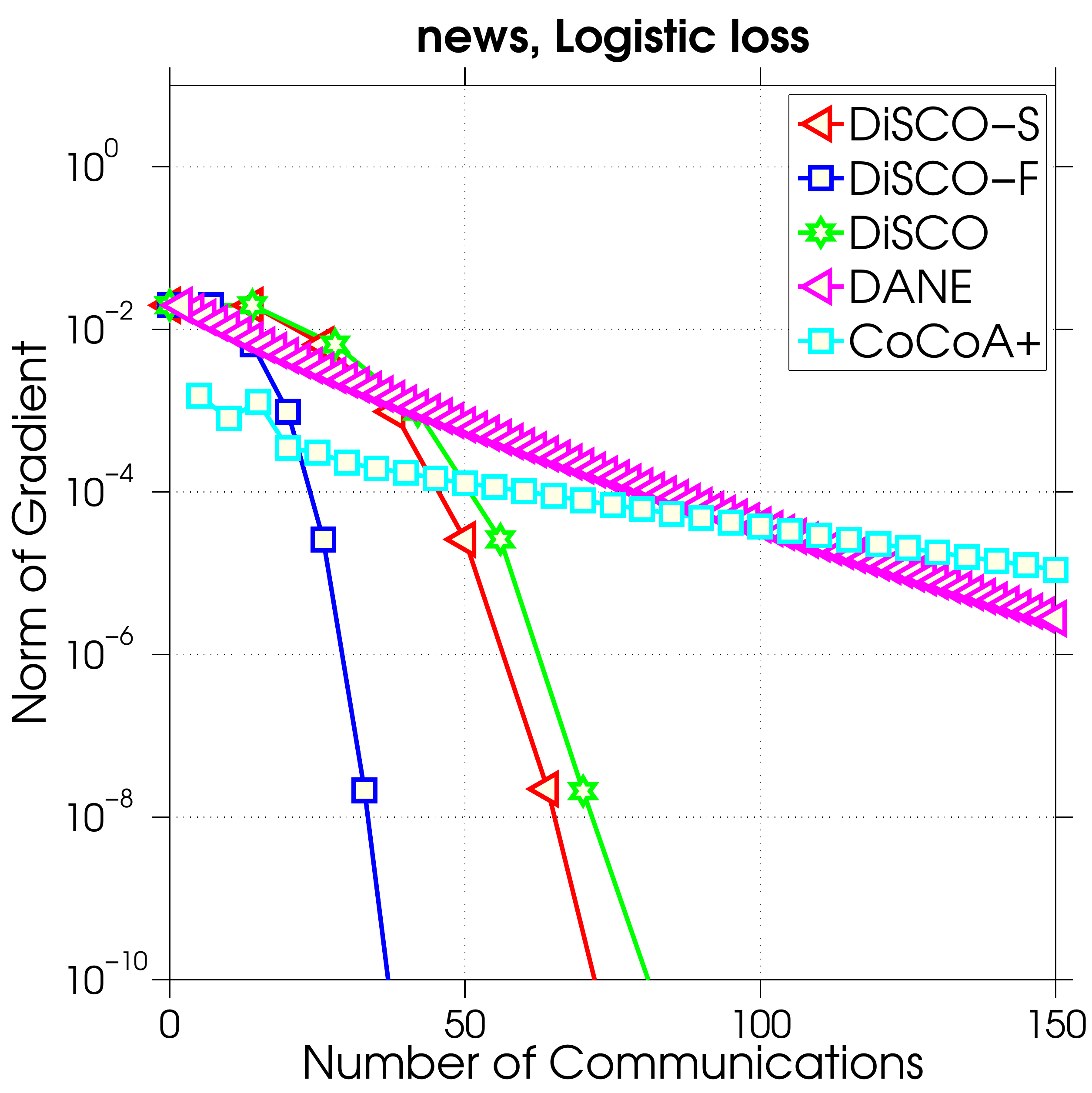}
\includegraphics[scale=.15]{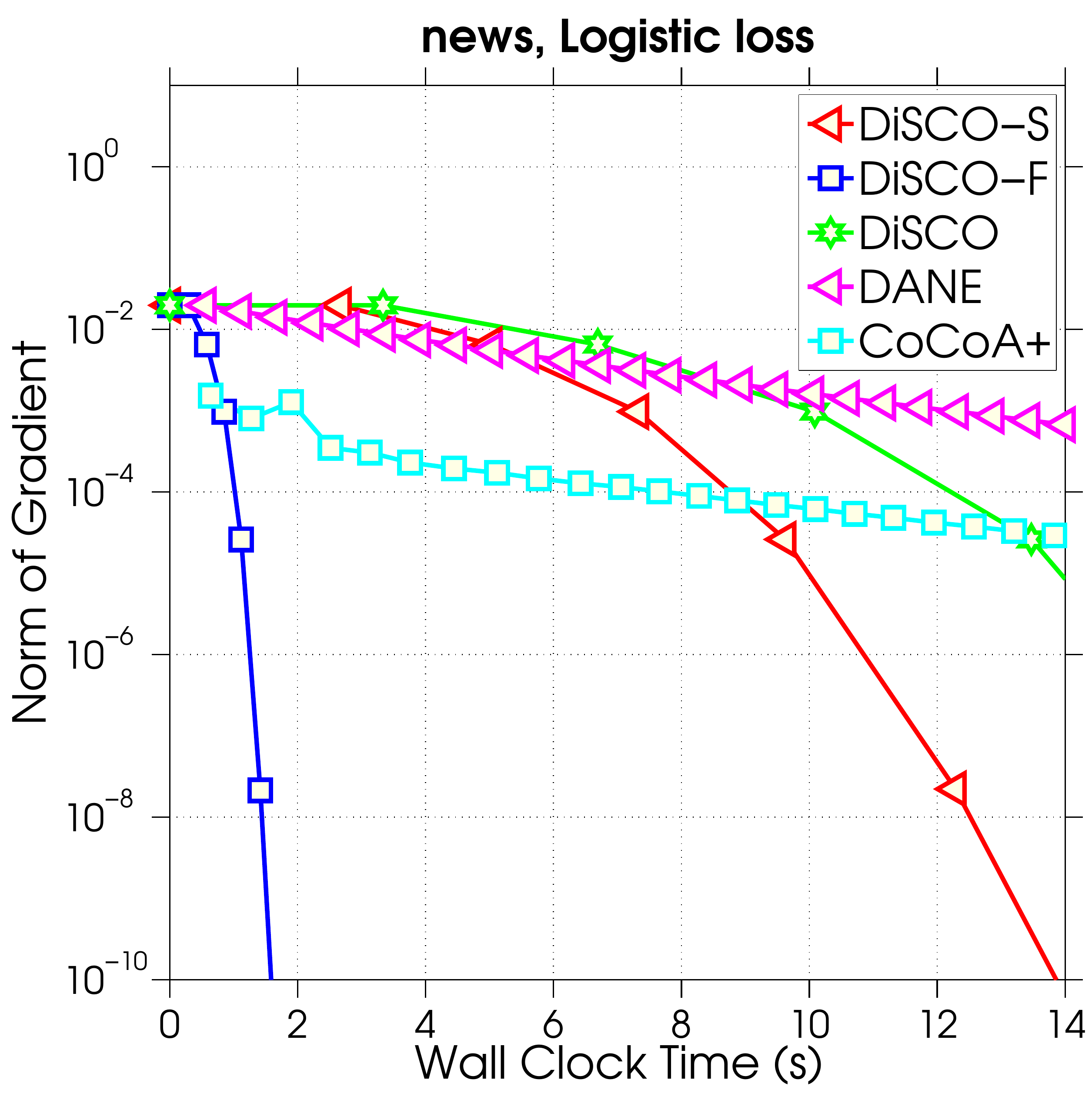}

\includegraphics[scale=.15]{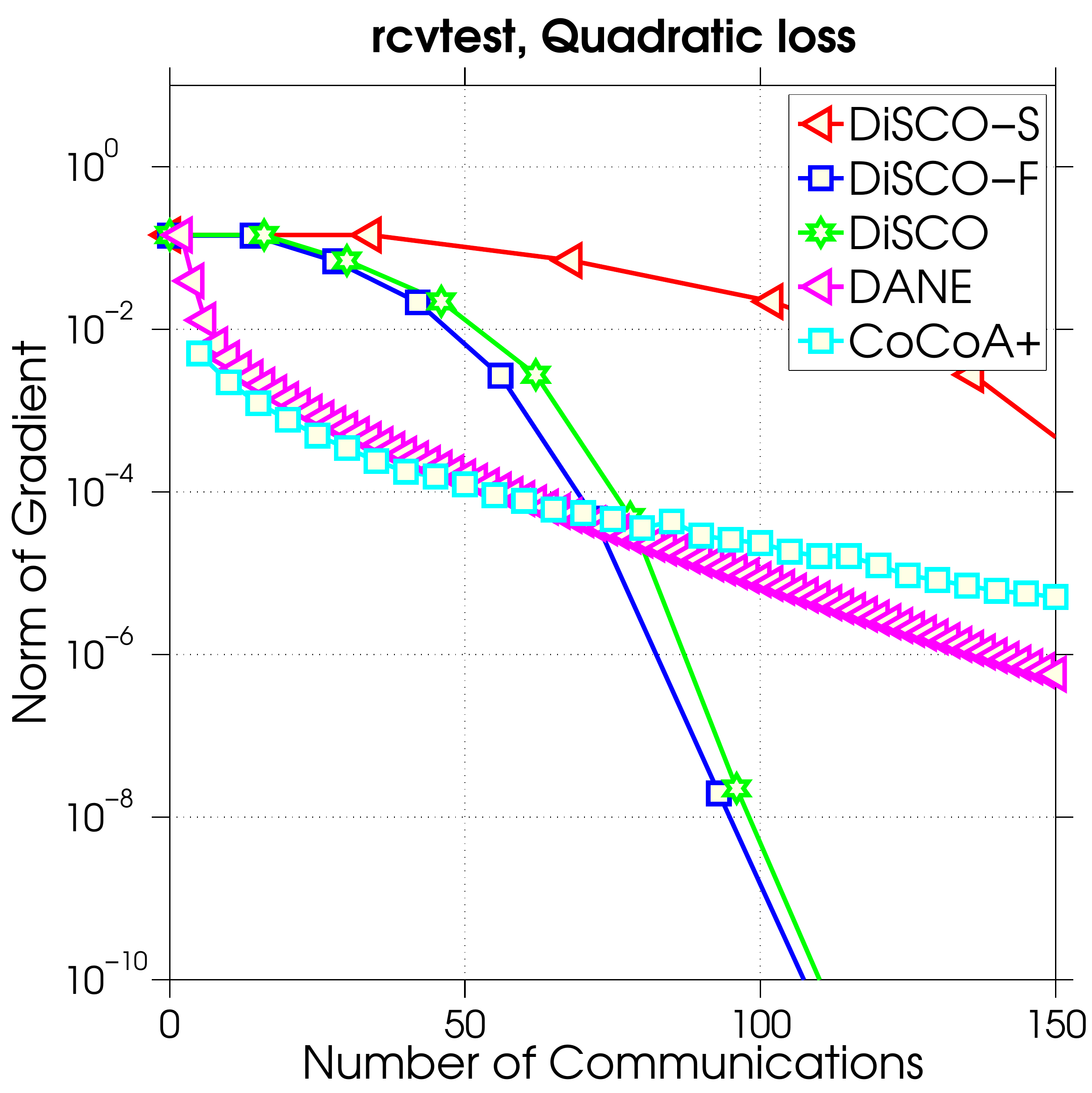}
\includegraphics[scale=.15]{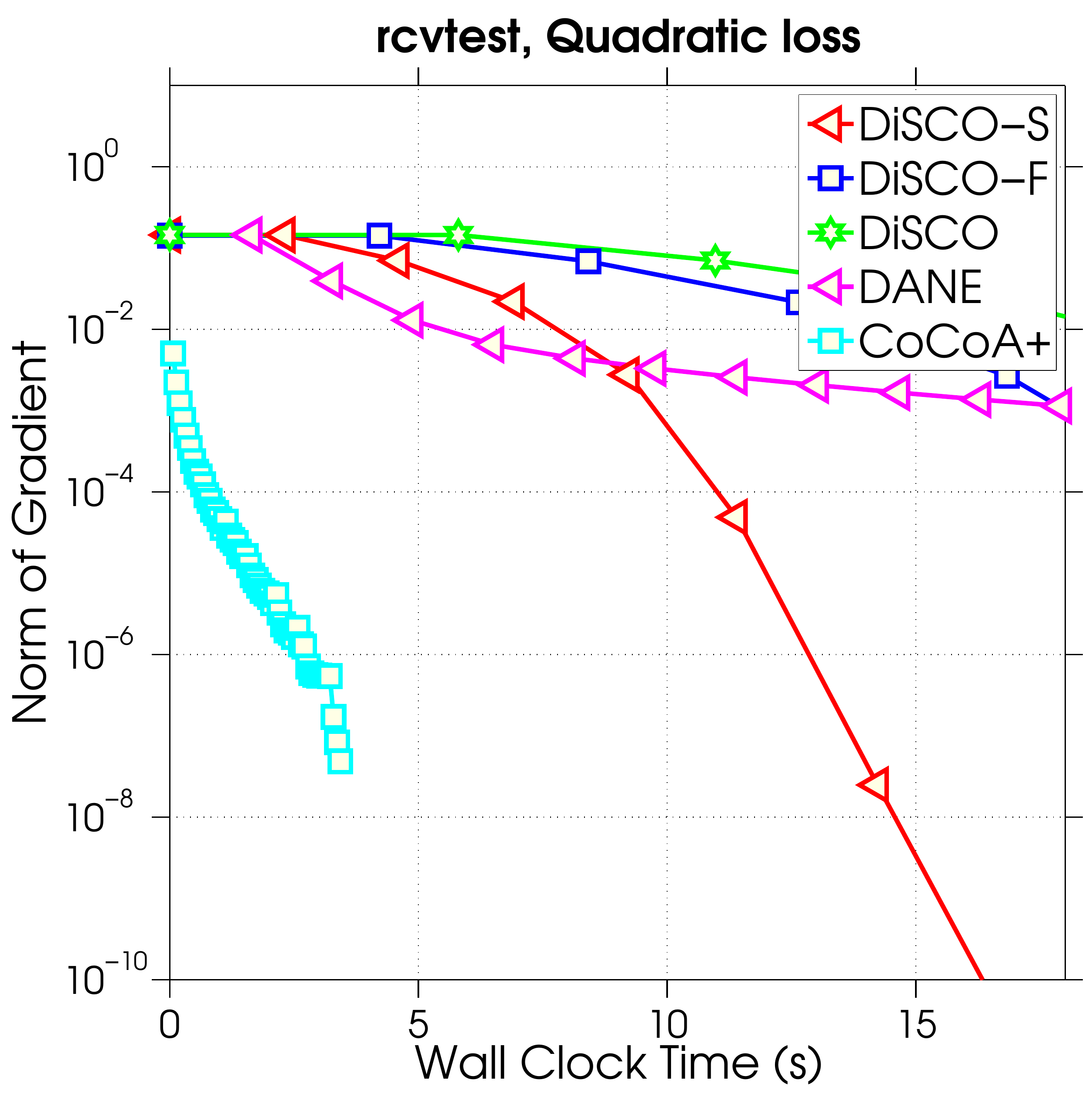}
\includegraphics[scale=.15]{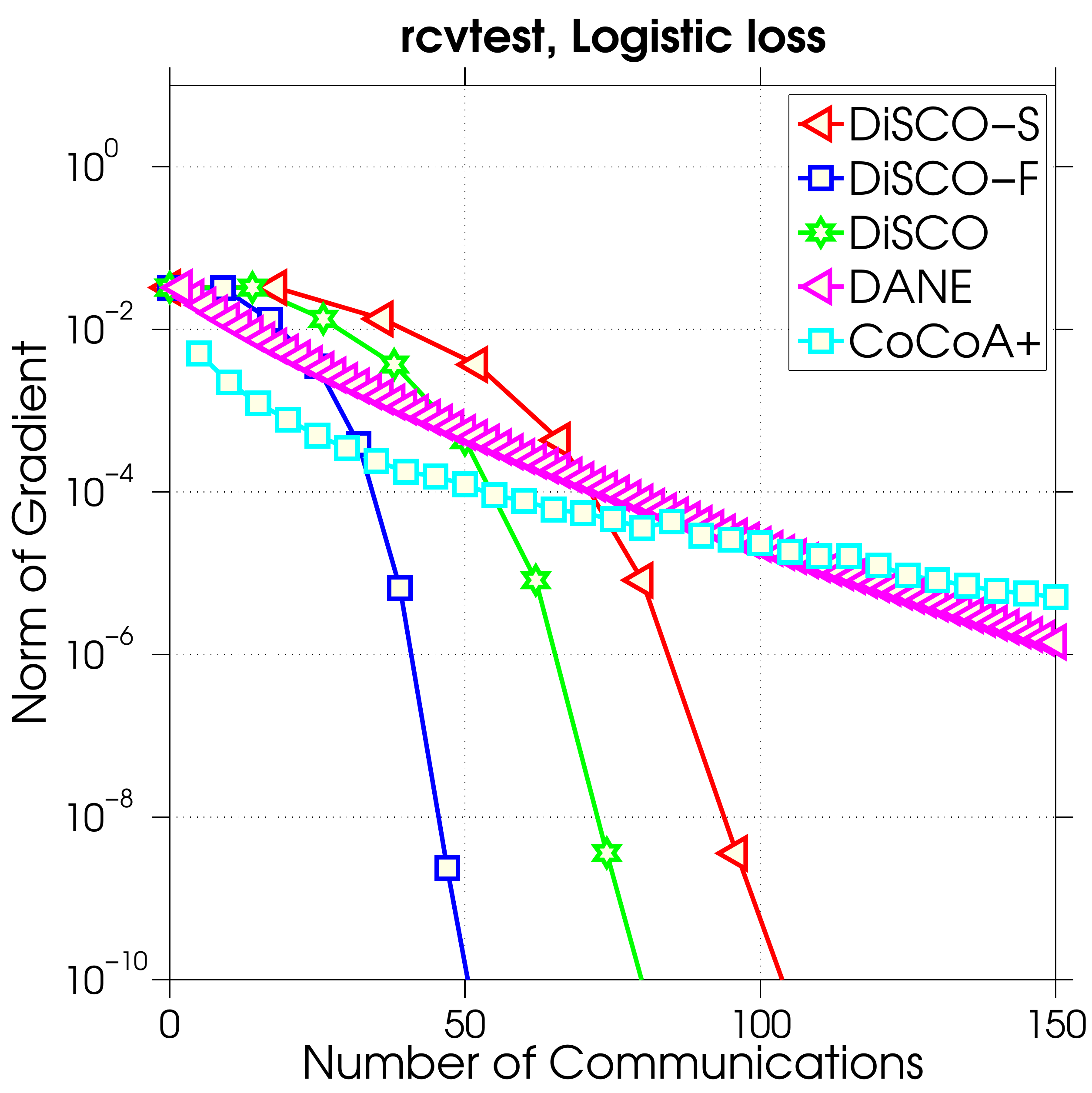}
\includegraphics[scale=.15]{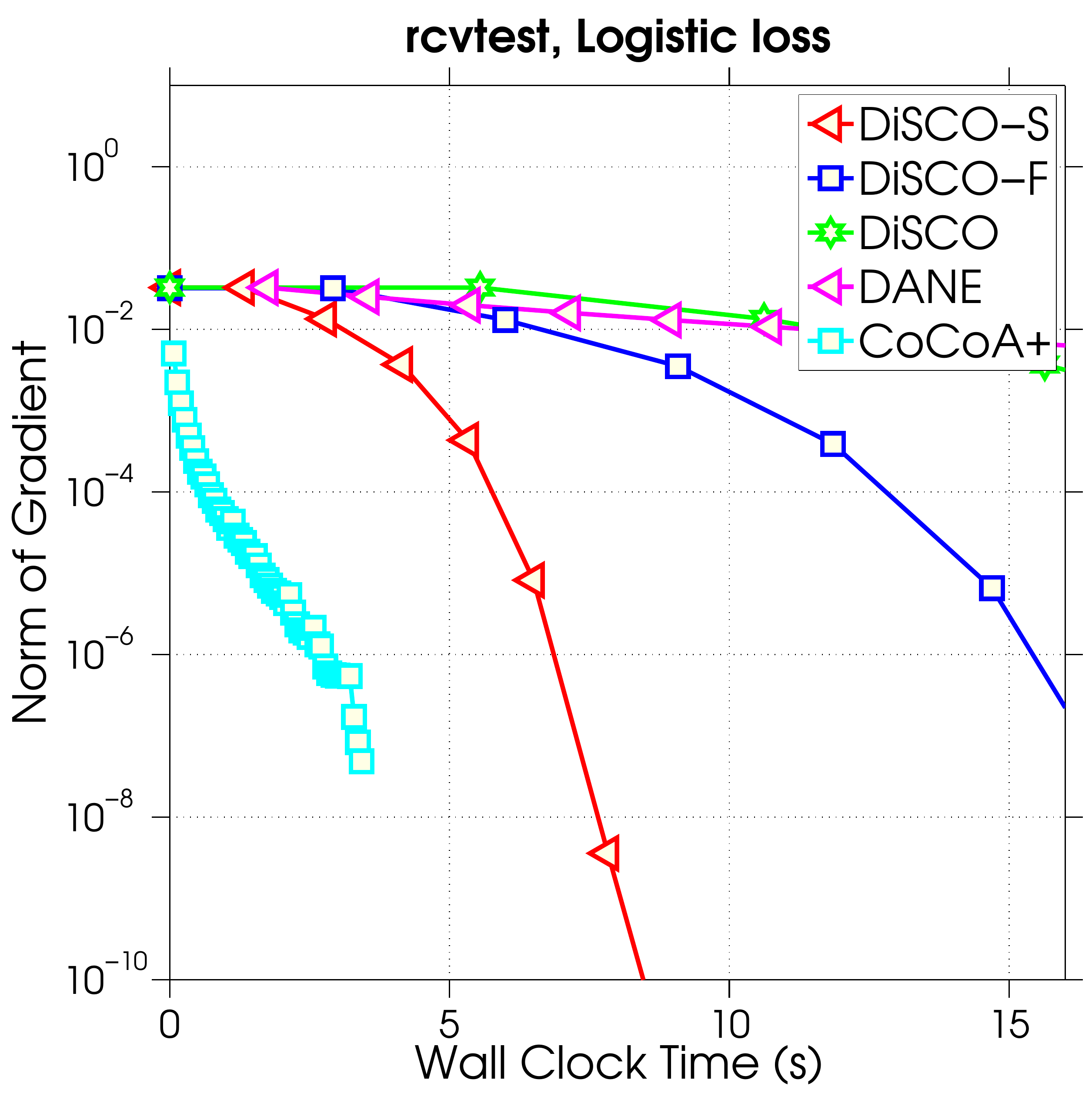}

\includegraphics[scale=.15]{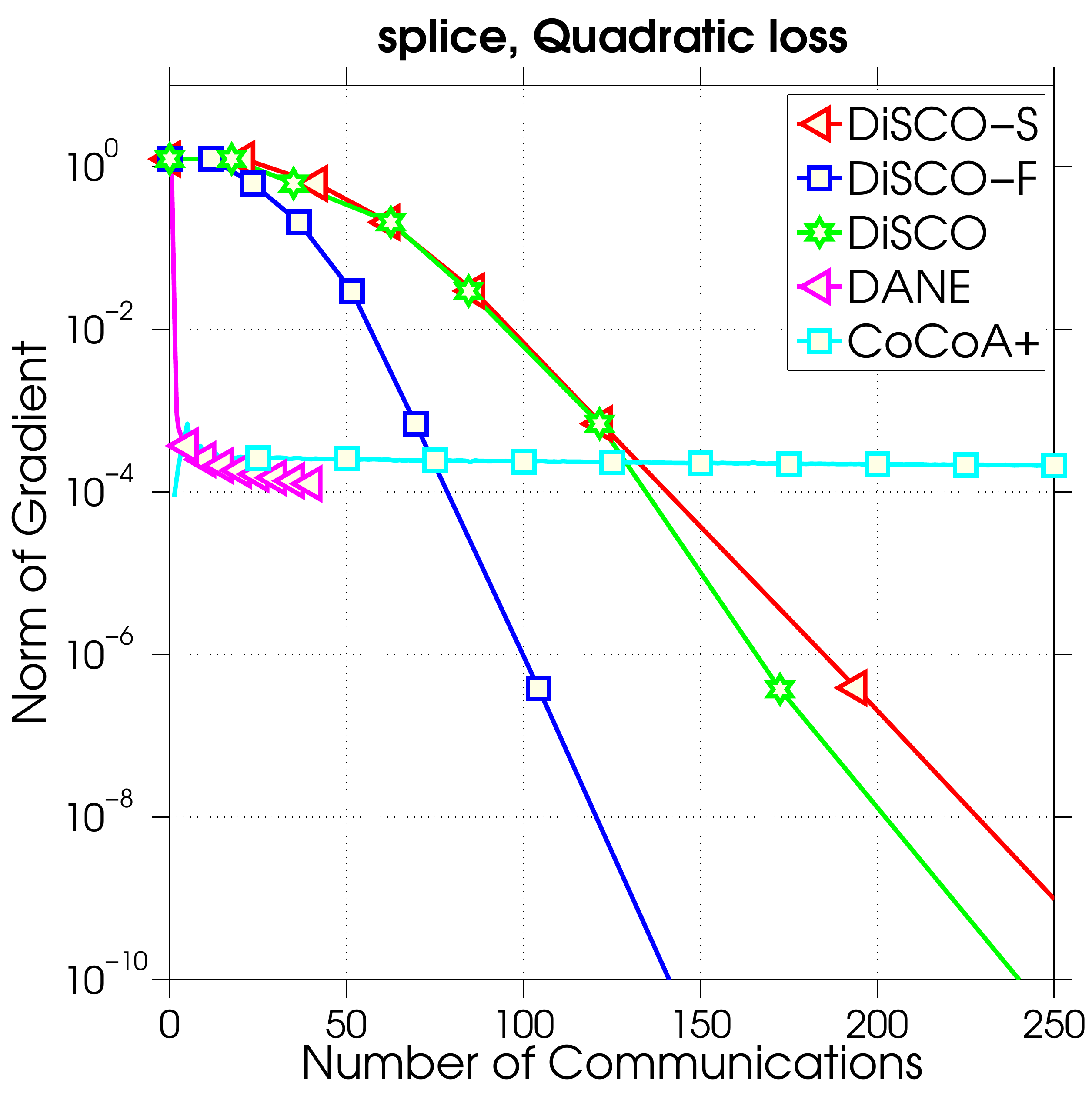}
\includegraphics[scale=.15]{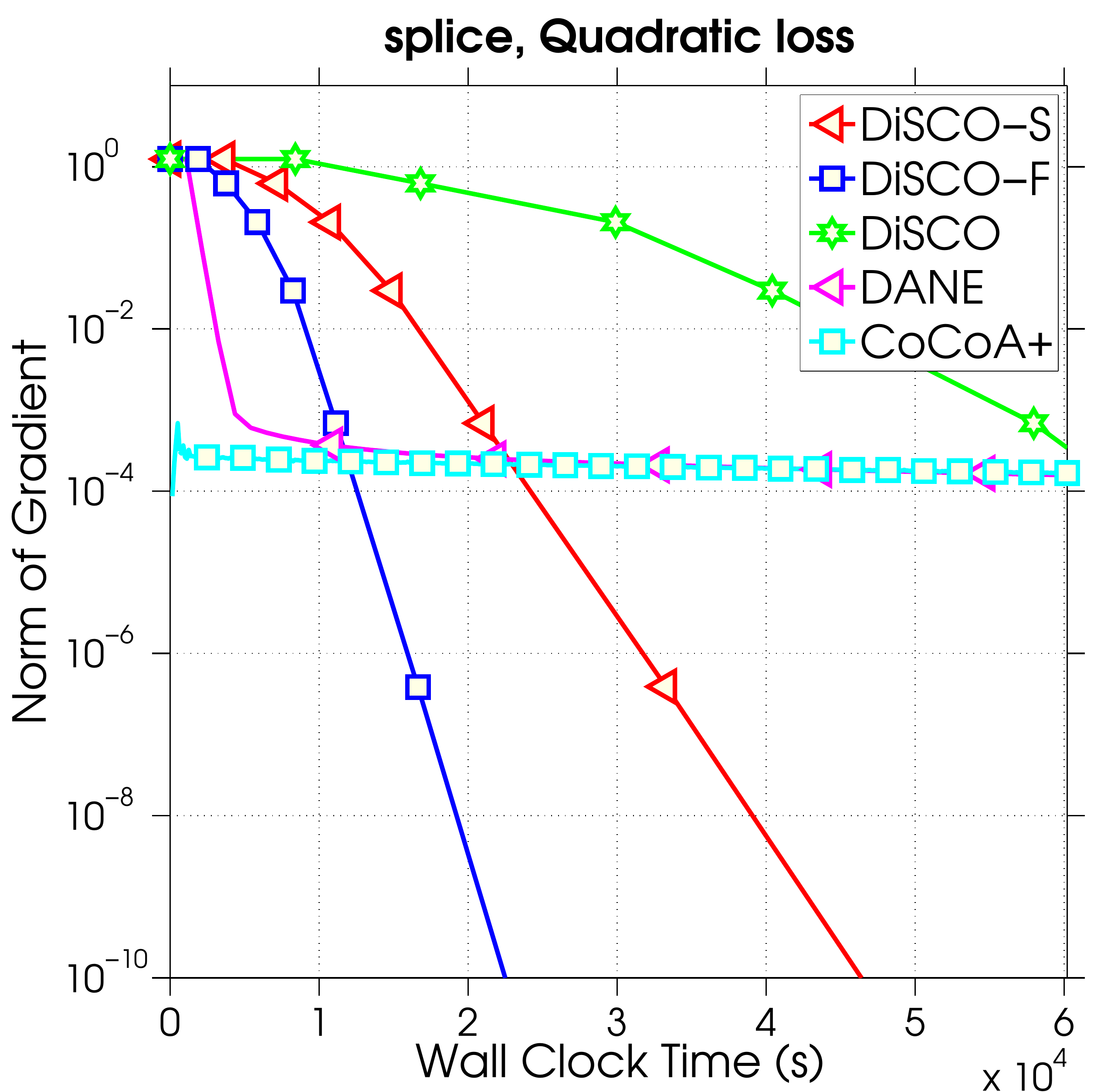}
\includegraphics[scale=.15]{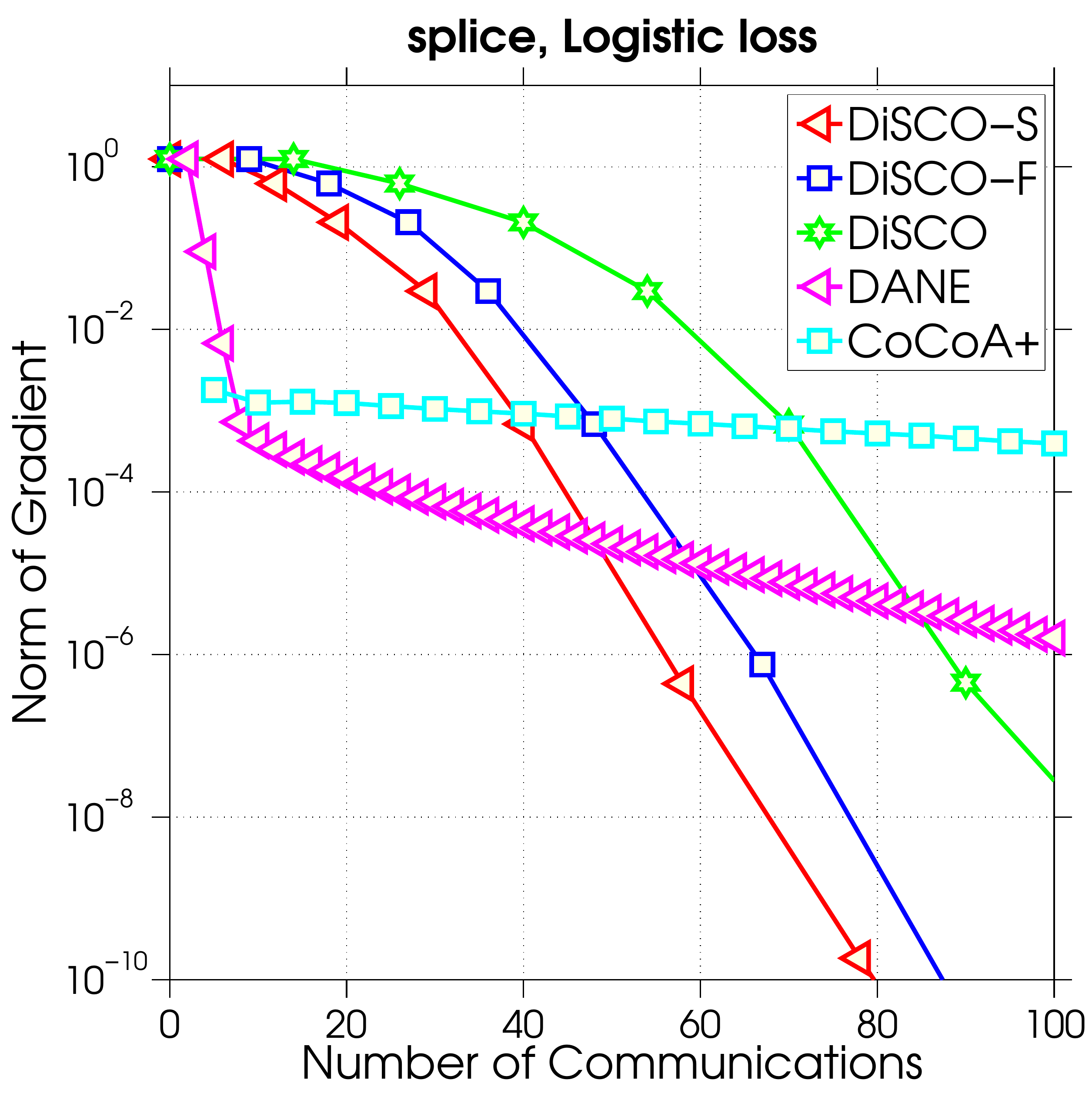}
\includegraphics[scale=.15]{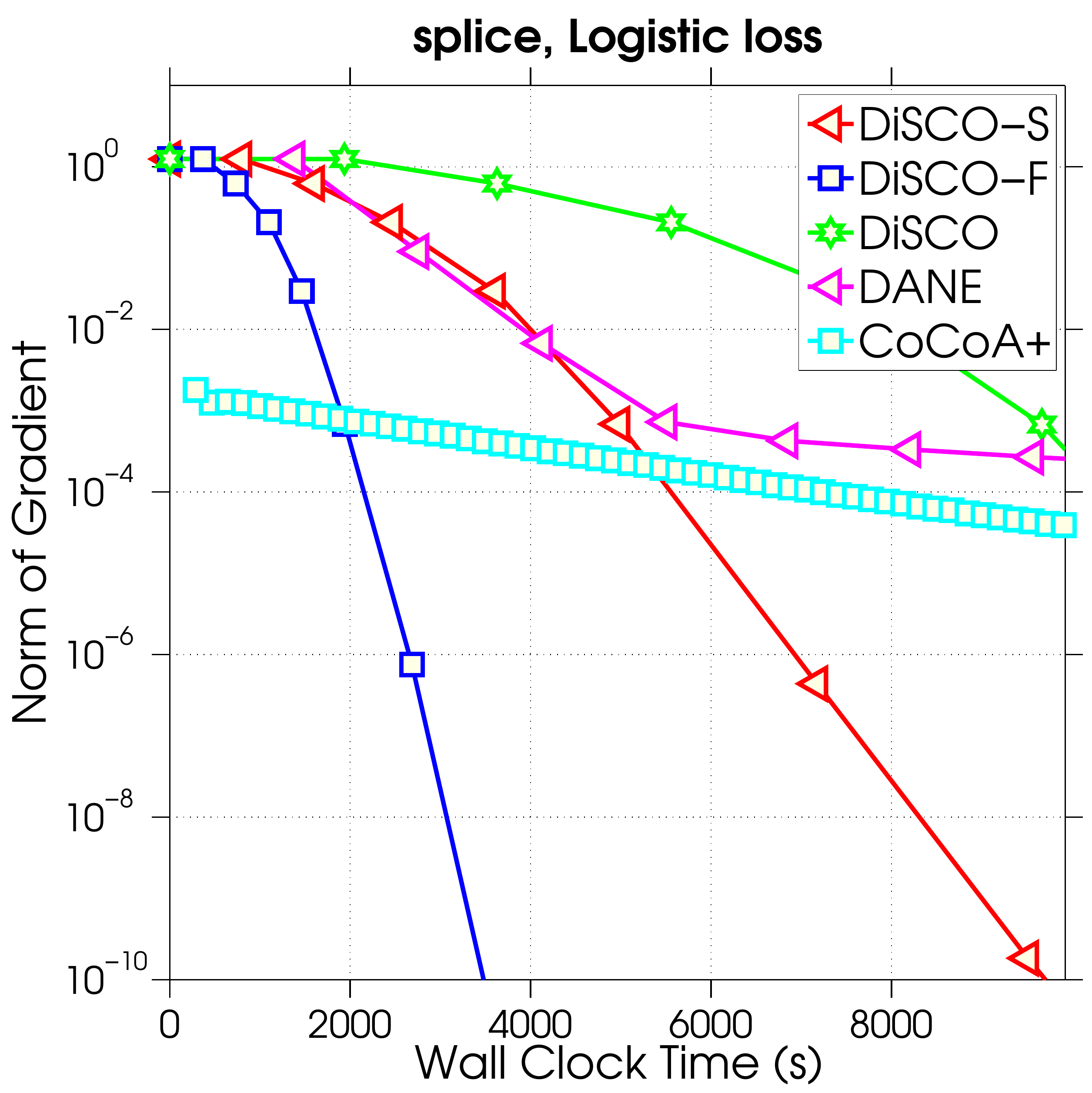}
\caption{Norm of gradient vs. the round of communication, as well as norm of gradient vs. elapsed time in seconds for three datasets: news20 (top, $\lambda = 1e-3$), rcv1.test (middle, $\lambda = 1e-4$), splice-site.test (bottom, $\lambda = 1e-6$) , each for two loss functions, Quadratic loss (left) and Logistic loss (right).}
\label{fig:dffsoll}
\end{figure*}

\subsection{Implementation Details}

We implement DiSCO and all other algorithms for comparison in C++, and run them in Amazon cluster using four m3.large EC2 instances. We apply all methods on applying Quadratic loss and Logistic loss in \eqref{eq:pro1}. A summary of the datasets used is shown in Table \ref{tab:datasets}.

 \begin{table}
\caption{Datasets used for numerical experiments.}
{
      \begin{tabular}{crrr}
      \hline
    Dataset & \multicolumn{1}{c}{$n$} &
    \multicolumn{1}{c}{$d$} & 
    \multicolumn{1}{c}{size(GB)} \\
    \hline 
  rcv1.test & 677,399 &
    47,236 & 1.21 \\
    news20 & 19,996 &
    1,355,191 & 0.13 \\
    splice-site.test & 4,627,840 &
    11,725,480 & 273.4
  \\  \hline  
      \end{tabular}
}    
\label{tab:datasets}   
\end{table}

\subsection{Comparison of different algorithms}

We compare the DiSCO-S, DiSCO-F, DiSCO, DANE and CoCOA+ directly using two datasets (new20 and rcv1.test) across two loss functions, where $\lambda$ is fixed to be $1e-4$. In DiSCO-S and DiSCO-F, we set $\tau = 100$. In DiSCO and DANE, we apply Stochastic Average Gradient(SAG) \cite{schmidt2013minimizing} to solve linear system $Ps = r$ and subproblem \eqref{eq:danesubp}, respectively. Also, $\mu$ was set as $1e-2$ for both of them. In CoCoA+, SDCA was used as the solver for subproblems. 

In Figure \ref{fig:dffsoll}, we plot how the norm of the gradient of objective decreases with respective to the total number of communication and the elapsed time. In all the cases, DiSCO-F uses only half of the rounds of communications compared with DiSCO-S. Also, DiSCO-S often uses similar rounds of communications with the original DiSCO, which demonstrates the advantage of using preconditioning matrix based on only a small subset of the samples. Finally, DANE and CoCoA+ will decrease the norm of gradient very fast at the first few iterations, but the decreasing become much weaker as the iterations continue.

For the news20 ($d\gg n$) and   splice-site.test ($d\sim n$) dataset, the DiSCO-F converges to the optimal solution with fewer iterations than all the other methods. The elapsed time for DiSCO-F is only 10\% of DiSCO-S in the news20 case, due to the smaller size of the vector that needs to be communicated. 

However, for the rcv.test dataset ($d< n$), even though DiSCO-F uses less number of communications, it tends to take longer time to reach an expected tolerance than DiSCO-S and CoCOA+.  This is because the longer vectors ($\R^n$) that DiSCO-F needs to communicate in each CG iteration, compared with them in DiSCO-S and CoCOA+ ($\R^d$). 

\subsection{Impact of the Parameter $\tau$}\label{sec:expP}

\begin{figure} 
\centering
\includegraphics[scale=.15]{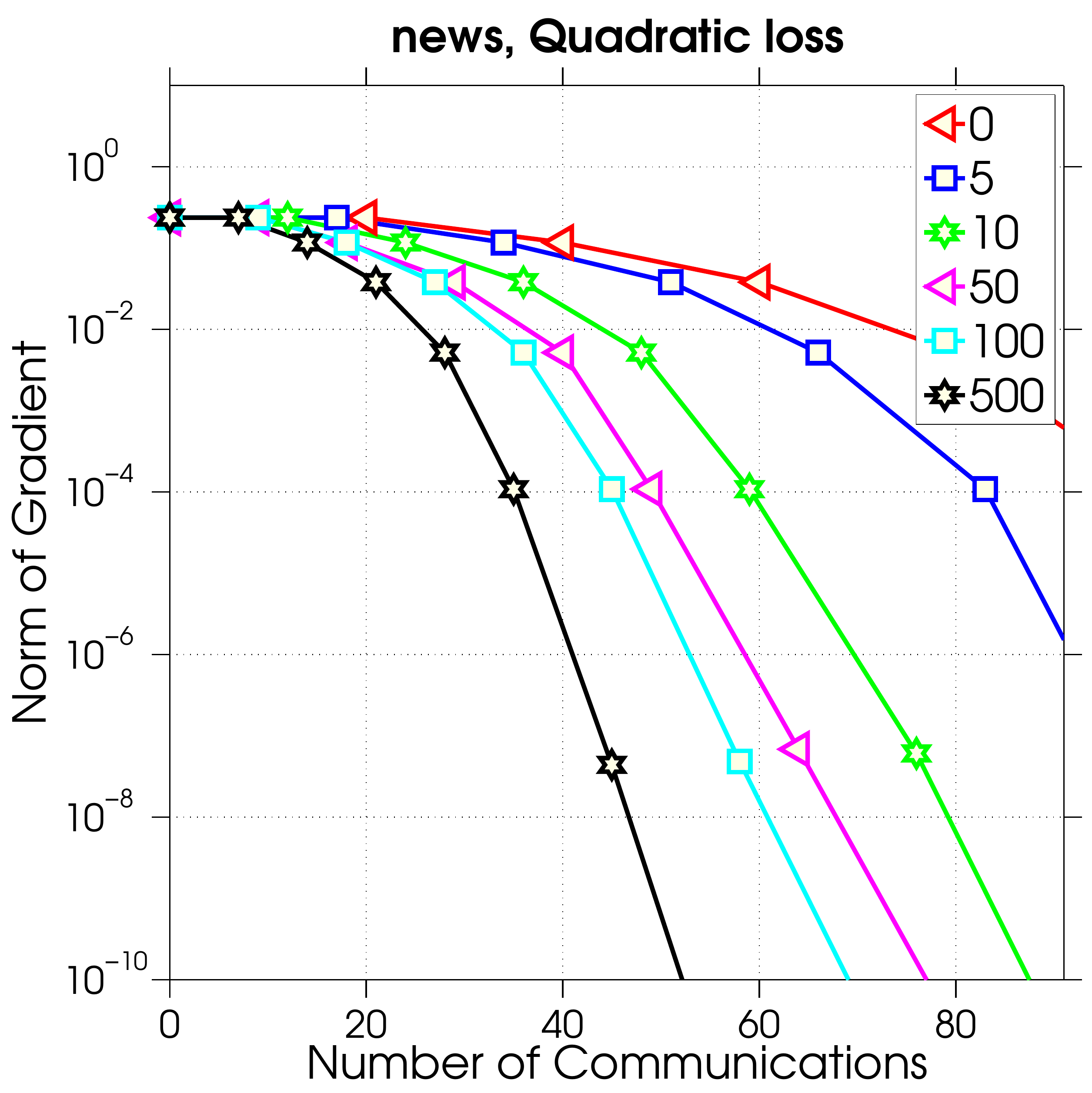}
\includegraphics[scale=.15]{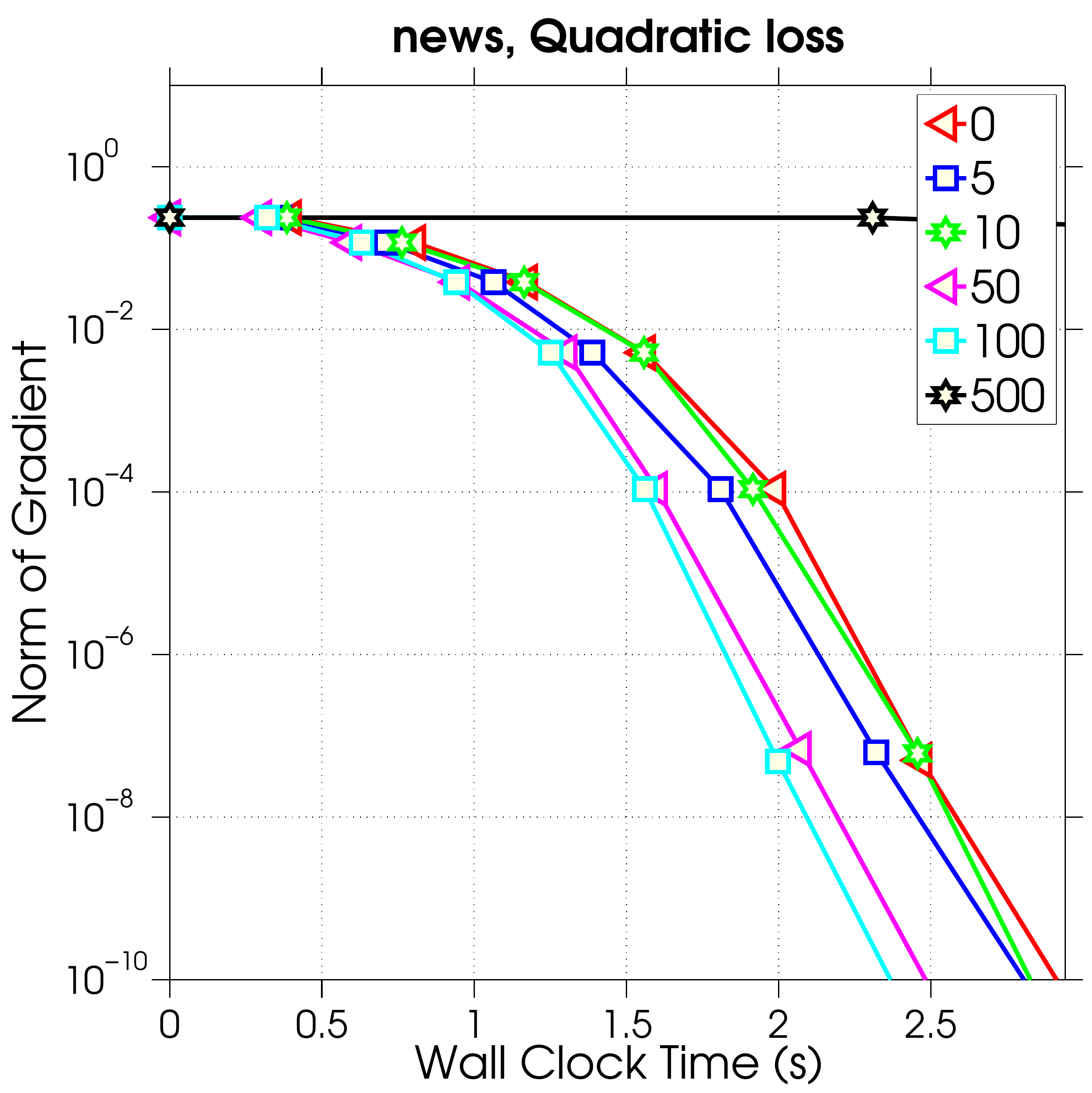}
\includegraphics[scale=.15]{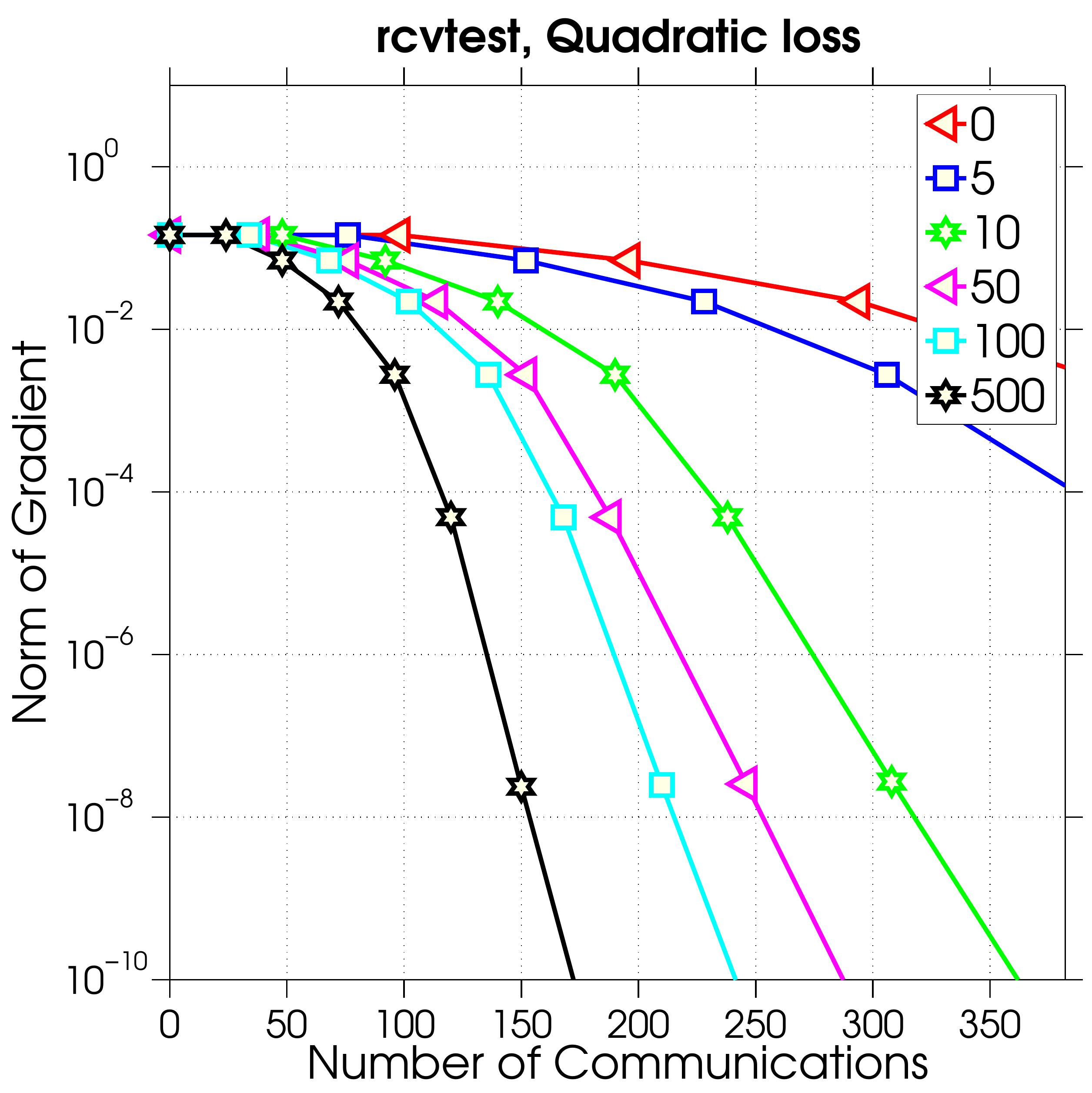}
\includegraphics[scale=.15]{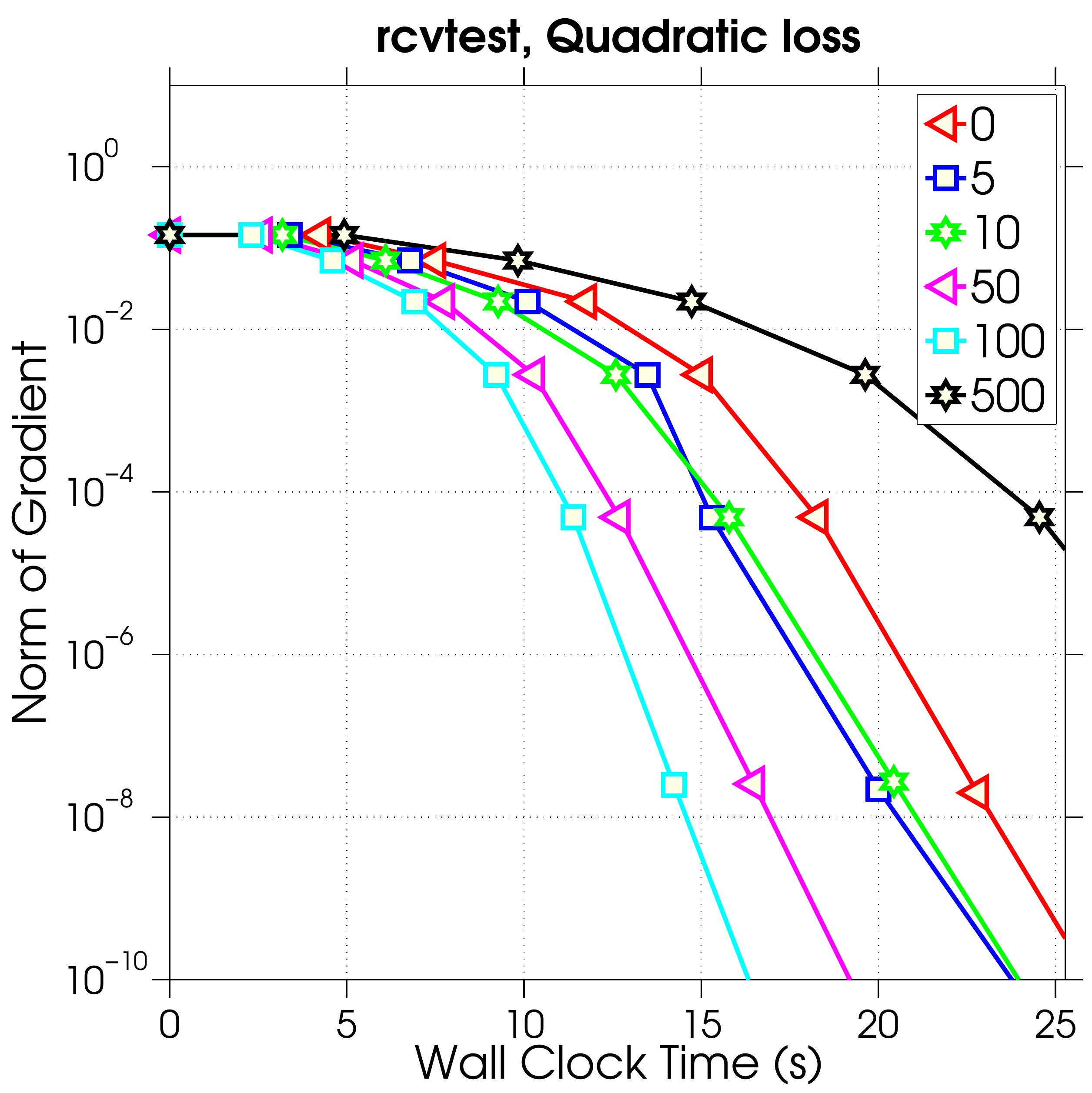}
\caption{Comparison of different number of samples used in Preconditioning by running DiSCO-F.} 
\label{fig:exppre}
\end{figure}
In this section, we compare the performance of DiSCO-F algorithm of setting different $\tau$. If we apply methods described in Section \ref{sec:wood}, the parameter $\tau$ would determine how well the preconditioning matrix $P$ can approximate the true Hessian $H$. In a extreme case, if we only use one machine and $\tau = n$, then $\|P-H\|_2 = 0$ and each iteration of Algorithm \ref{Disco} will only use 1 iteration of CG algorithm. However, too large $\tau$ will cause computation in Algorithm \ref{A2222} quite expensive, thus resulting in long elapsed time. In our experiment, $\tau = 500$ is even not acceptable, in terms of elapsed time.

As shown in Figure \ref{fig:exppre}, the larger $\tau$ we use, the less total number of communications the algorithm takes to reach optimality. However, $\tau=100$ always leads to shorter time in both of these two datasets.

\subsection{How Many Samples to Compute Hessian?}\label{sec:expHess}

\begin{figure} 
\centering
\includegraphics[scale=.15]{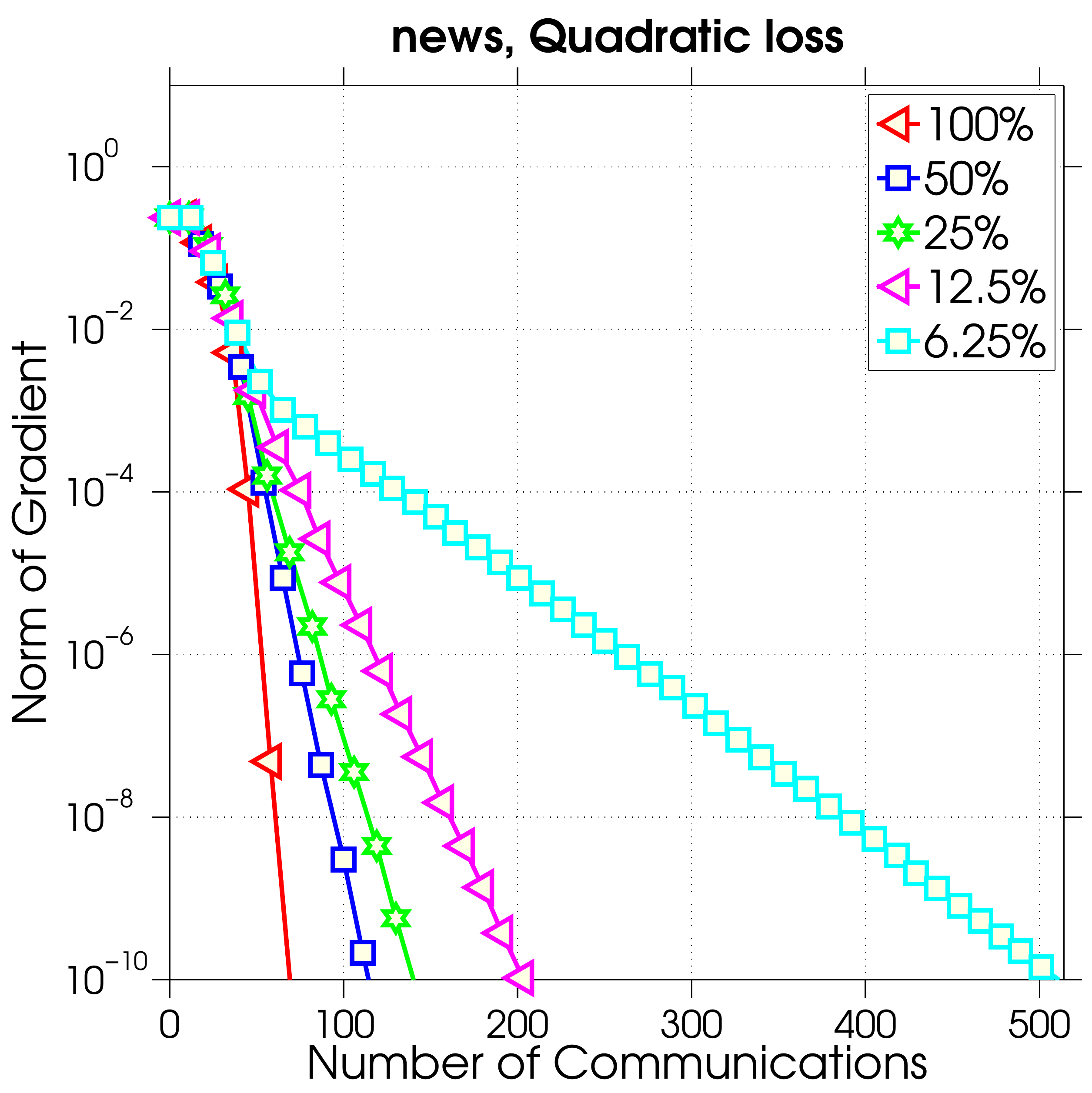}
\includegraphics[scale=.15]{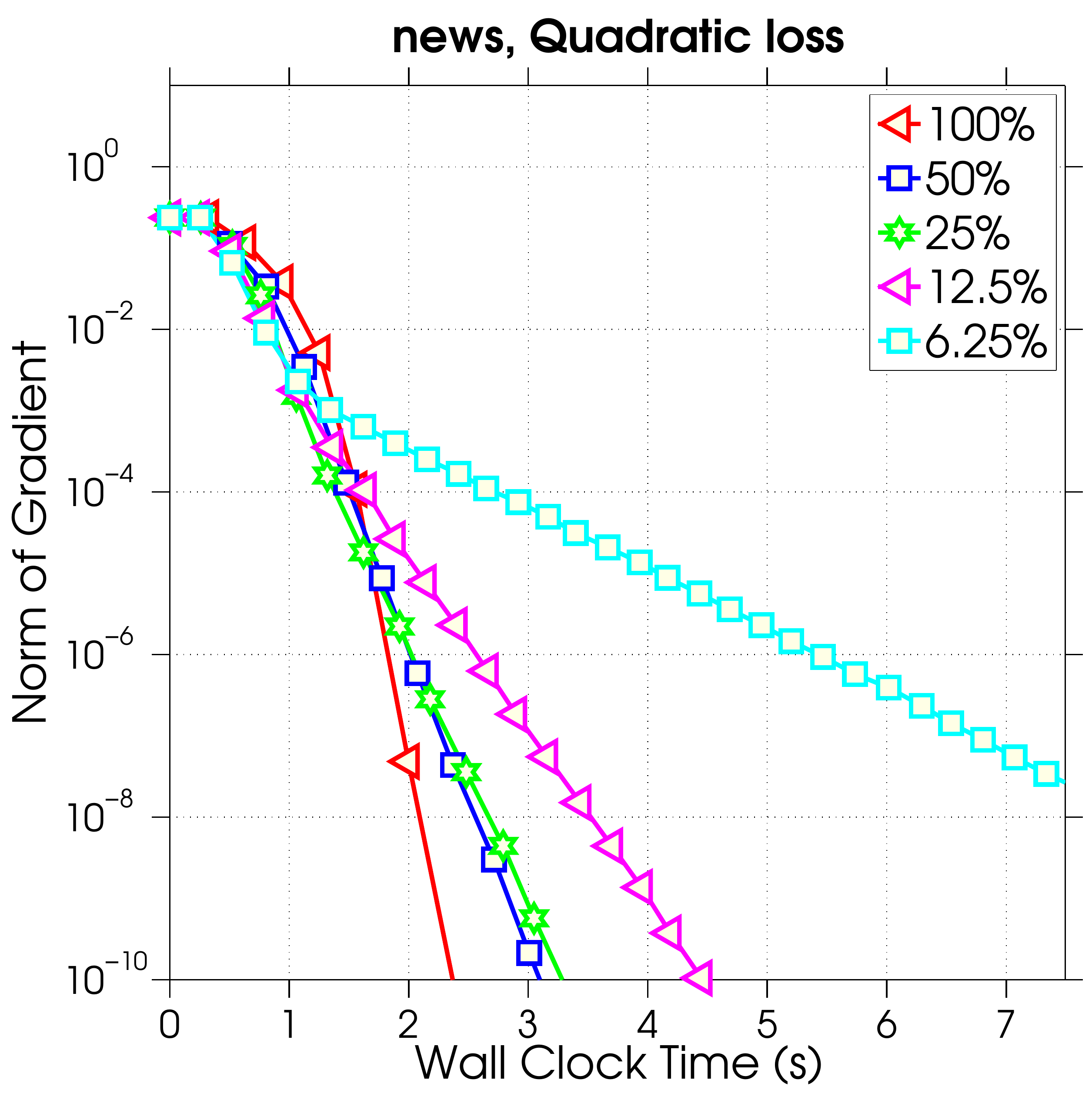}
\includegraphics[scale=.15]{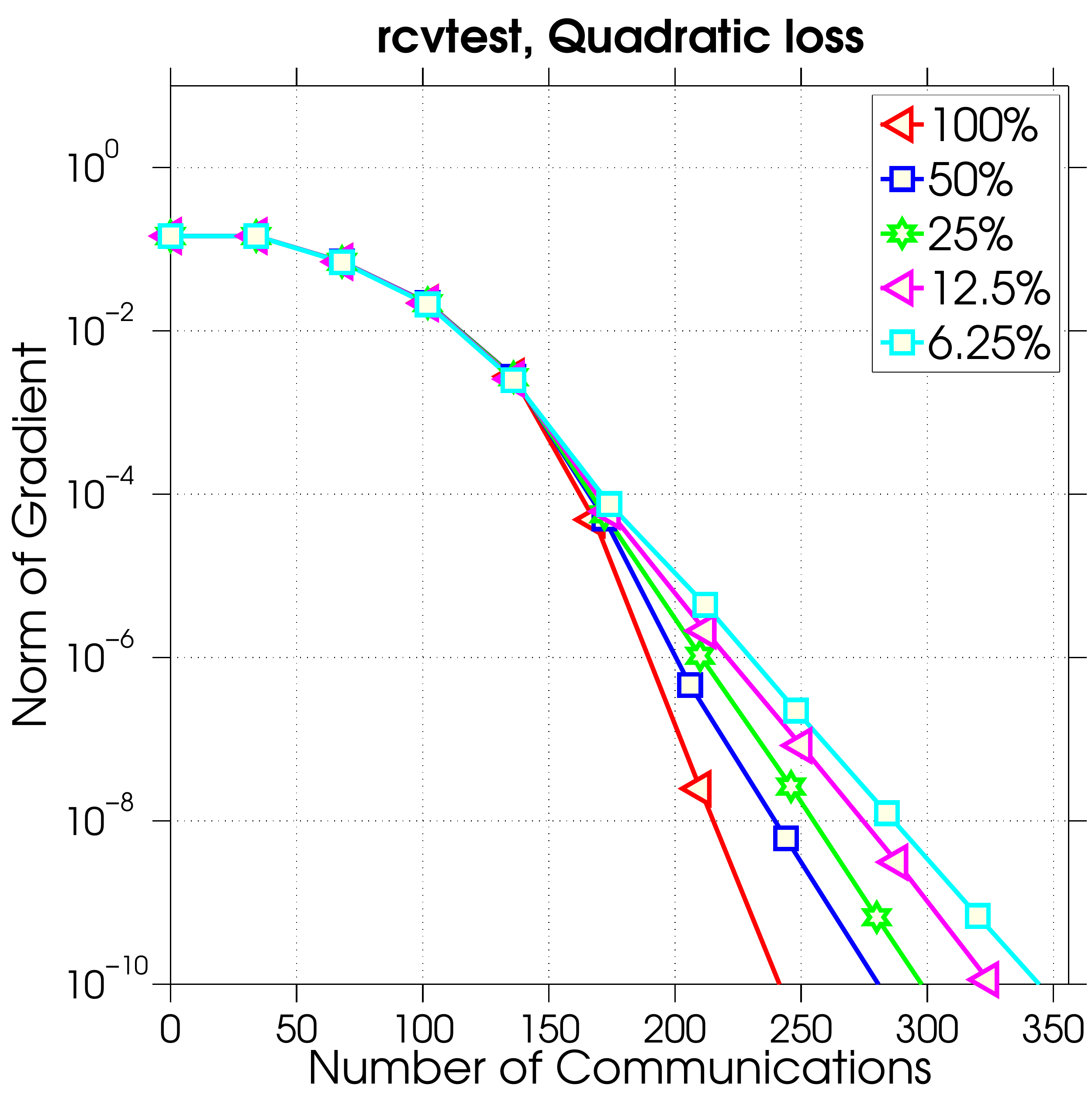}
\includegraphics[scale=.15]{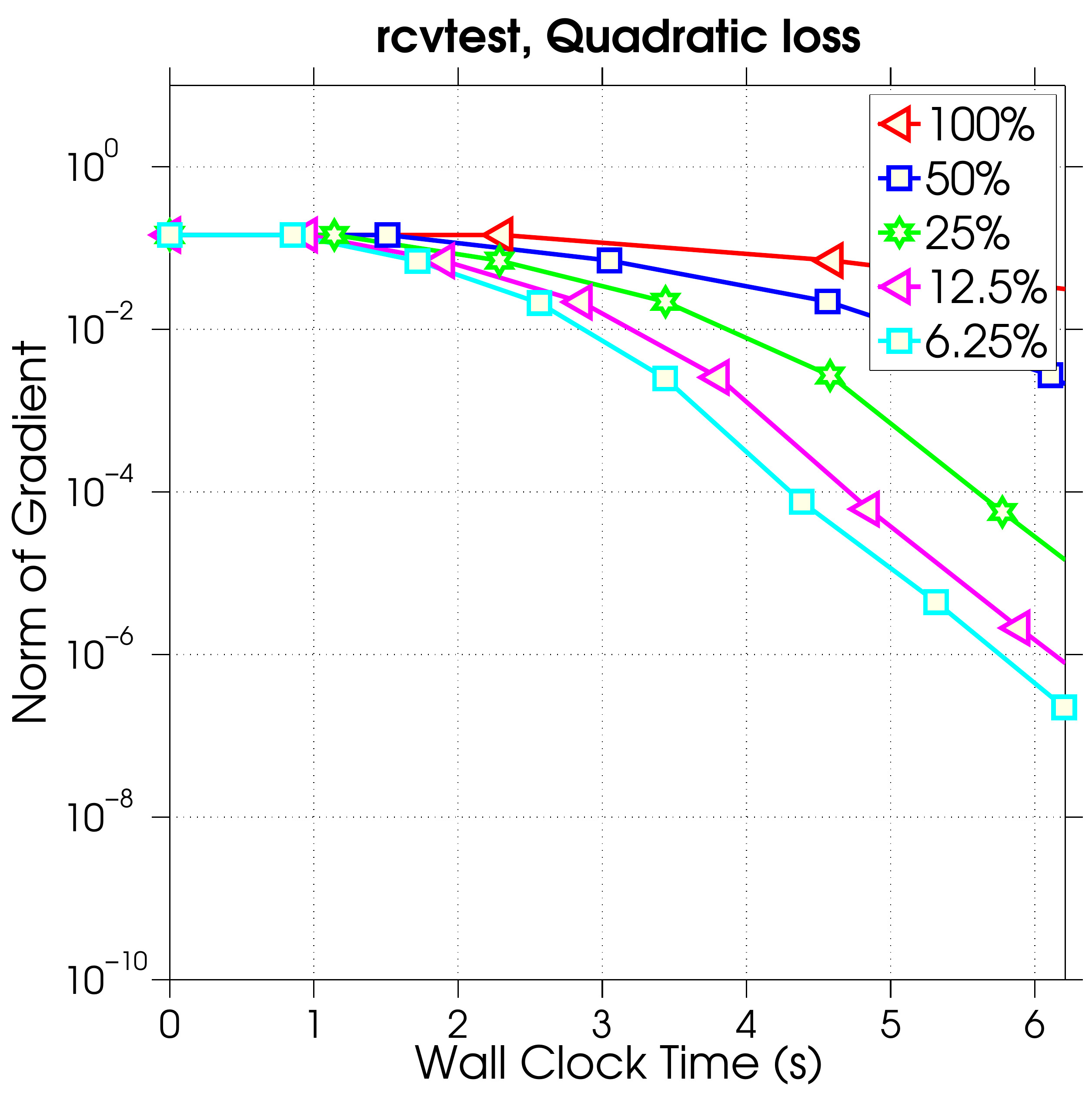}
\caption{Comparison of different number of samples used in approximating Hessian by running DiSCO-F.} 
\label{fig:Hessian}
\end{figure}

Notice that in step 4 of Algorithm \ref{A1} and \ref{A2}, the product of the Hessian matrix and a vector need to be computed. In this section, We would like to try reducing the size of samples to compute Hessian. By doing so, we have to give up the current guaranteed complexity, since now the Hessian will be approximated. However, less elapsed time is expected if we choose proper size of samples. Due to the lack of theoretical analysis on this attempt, we only list the observation from the experiments. 

For each iteration of Algorithm \ref{Disco}, we choose a subset of samples uniformly randomly to get the approximated Hessian.  We try to choose subsets of samples whose sizes range from 100\% to 6.25\% of the entire dataset, as shown in Figure \ref{fig:Hessian}. For news20 dataset, such attempt will bring no benefits. The more samples we use, the less round of communications and elapsed time the algorithm spends to reach optimality. But for rcv1.test dataset, the elapsed time decreases as we reduce the number of samples to compute Hessian, which illustrates that using only a small portion of samples will be helpful to get enough information of Hessian in each iteration.

The reason of this result might be for the dataset with a rather large number of features (news20), ignoring some samples will result in lots of relationship between features missed. For the dataset with rather small number of features (rcv1), the Hessian can be approximated well by only a small subset of data.

\section{Conclusion}

In conclusion, based on the DiSCO algorithm \cite{zhang2015communication}, we study inexact dumped Newton method implemented in a distributed way. We found that by partitioning the dataset by features, the number of communications can be reduced and the computation in each machine becomes more balanced. Also, we shrink the size of samples to generate the preconditioning matrix, which greatly improves the efficiency of solving the linear system in CG. Our experimental results show significant speedups over previous methods, including the original DiSCO algorithm as well as other state-of-the-art methods.


\bibliography{literature}

\begin{thebibliography}{36}
\providecommand{\natexlab}[1]{#1}
\providecommand{\url}[1]{\texttt{#1}}
\expandafter\ifx\csname urlstyle\endcsname\relax
  \providecommand{\doi}[1]{doi: #1}\else
  \providecommand{\doi}{doi: \begingroup \urlstyle{rm}\Url}\fi

\bibitem[Agarwal \& Duchi(2011)Agarwal and Duchi]{agarwal2011distributed}
Agarwal, Alekh and Duchi, John~C.
\newblock Distributed delayed stochastic optimization.
\newblock In \emph{Advances in Neural Information Processing Systems}, pp.\
  873--881, 2011.

\bibitem[Bertsekas \& Tsitsiklis(1989)Bertsekas and
  Tsitsiklis]{bertsekas1989parallel}
Bertsekas, Dimitri~P and Tsitsiklis, John~N.
\newblock \emph{Parallel and distributed computation: numerical methods}.
\newblock Prentice-Hall, Inc., 1989.

\bibitem[Boyd et~al.(2011)Boyd, Parikh, Chu, Peleato, and
  Eckstein]{boyd2011distributed}
Boyd, Stephen, Parikh, Neal, Chu, Eric, Peleato, Borja, and Eckstein, Jonathan.
\newblock Distributed optimization and statistical learning via the alternating
  direction method of multipliers.
\newblock \emph{Foundations and Trends{\textregistered} in Machine Learning},
  3\penalty0 (1):\penalty0 1--122, 2011.

\bibitem[Bradley et~al.(2011)Bradley, Kyrola, Bickson, and
  Guestrin]{bradley2011parallel}
Bradley, Joseph~K, Kyrola, Aapo, Bickson, Danny, and Guestrin, Carlos.
\newblock Parallel coordinate descent for l1-regularized loss minimization.
\newblock \emph{arXiv preprint arXiv:1105.5379}, 2011.

\bibitem[Csiba et~al.(2015)Csiba, Qu, and Richt{\'a}rik]{csiba2015stochastic}
Csiba, Dominik, Qu, Zheng, and Richt{\'a}rik, Peter.
\newblock Stochastic dual coordinate ascent with adaptive probabilities.
\newblock \emph{arXiv preprint arXiv:1502.08053}, 2015.

\bibitem[Defazio et~al.(2014)Defazio, Bach, and
  Lacoste-Julien]{defazio2014saga}
Defazio, Aaron, Bach, Francis, and Lacoste-Julien, Simon.
\newblock Saga: A fast incremental gradient method with support for
  non-strongly convex composite objectives.
\newblock In \emph{Advances in Neural Information Processing Systems}, pp.\
  1646--1654, 2014.

\bibitem[Dekel et~al.(2012)Dekel, Gilad-Bachrach, Shamir, and
  Xiao]{dekel2012optimal}
Dekel, Ofer, Gilad-Bachrach, Ran, Shamir, Ohad, and Xiao, Lin.
\newblock Optimal distributed online prediction using mini-batches.
\newblock \emph{The Journal of Machine Learning Research}, 13\penalty0
  (1):\penalty0 165--202, 2012.

\bibitem[Deng \& Yin(2012)Deng and Yin]{deng2012global}
Deng, Wei and Yin, Wotao.
\newblock On the global and linear convergence of the generalized alternating
  direction method of multipliers.
\newblock \emph{Journal of Scientific Computing}, pp.\  1--28, 2012.

\bibitem[Duchi et~al.(2013)Duchi, Jordan, and McMahan]{Duchi:2013te}
Duchi, John~C, Jordan, Michael~I, and McMahan, H~Brendan.
\newblock {Estimation, Optimization, and Parallelism when Data is Sparse}.
\newblock In \emph{NIPS}, 2013.

\bibitem[Forero et~al.(2010)Forero, Cano, and Giannakis]{forero2010consensus}
Forero, Pedro~A, Cano, Alfonso, and Giannakis, Georgios~B.
\newblock Consensus-based distributed support vector machines.
\newblock \emph{The Journal of Machine Learning Research}, 11:\penalty0
  1663--1707, 2010.

\bibitem[Hsieh et~al.(2008)Hsieh, Chang, Lin, Keerthi, and
  Sundararajan]{hsieh2008dual}
Hsieh, Cho-Jui, Chang, Kai-Wei, Lin, Chih-Jen, Keerthi, S~Sathiya, and
  Sundararajan, Sellamanickam.
\newblock A dual coordinate descent method for large-scale linear svm.
\newblock In \emph{Proceedings of the 25th international conference on Machine
  learning}, pp.\  408--415. ACM, 2008.

\bibitem[Jaggi et~al.(2014)Jaggi, Smith, Tak{\'a}c, Terhorst, Krishnan,
  Hofmann, and Jordan]{jaggi2014communication}
Jaggi, Martin, Smith, Virginia, Tak{\'a}c, Martin, Terhorst, Jonathan,
  Krishnan, Sanjay, Hofmann, Thomas, and Jordan, Michael~I.
\newblock Communication-efficient distributed dual coordinate ascent.
\newblock In \emph{Advances in Neural Information Processing Systems}, pp.\
  3068--3076, 2014.

\bibitem[Johnson \& Zhang(2013)Johnson and Zhang]{johnson2013accelerating}
Johnson, Rie and Zhang, Tong.
\newblock Accelerating stochastic gradient descent using predictive variance
  reduction.
\newblock \emph{NIPS}, pp.\  315--323, 2013.

\bibitem[Kone{\v{c}}n{\'y} et~al.(2014)Kone{\v{c}}n{\'y}, Liu, Richt{\'a}rik,
  and Tak{\'a}{\v{c}}]{konevcny2014ms2gd}
Kone{\v{c}}n{\'y}, Jakub, Liu, Jie, Richt{\'a}rik, Peter, and Tak{\'a}{\v{c}},
  Martin.
\newblock {mS2GD:} {M}ini-batch semi-stochastic gradient descent in the
  proximal setting.
\newblock \emph{arXiv preprint arXiv:1410.4744}, 2014.

\bibitem[Lee \& Roth(2015)Lee and Roth]{lee2015distributed}
Lee, Ching-Pei and Roth, Dan.
\newblock Distributed box-constrained quadratic optimization for dual linear
  {SVM}.
\newblock ICML, 2015.

\bibitem[Liu et~al.(2014)Liu, Wright, R{\'e}, Bittorf, and Sridhar]{Liu:2014wj}
Liu, Ji, Wright, Stephen~J, R{\'e}, Christopher, Bittorf, Victor, and Sridhar,
  Srikrishna.
\newblock {An Asynchronous Parallel Stochastic Coordinate Descent Algorithm}.
\newblock In \emph{ICML}, 2014.

\bibitem[Ma et~al.(2015{\natexlab{a}})Ma, Kone{\v{c}}n{\`y}, Jaggi, Smith,
  Jordan, Richt{\'a}rik, and Tak{\'a}{\v{c}}]{ma2015distributed}
Ma, Chenxin, Kone{\v{c}}n{\`y}, Jakub, Jaggi, Martin, Smith, Virginia, Jordan,
  Michael~I, Richt{\'a}rik, Peter, and Tak{\'a}{\v{c}}, Martin.
\newblock Distributed optimization with arbitrary local solvers.
\newblock \emph{arXiv preprint arXiv:1512.04039}, 2015{\natexlab{a}}.

\bibitem[Ma et~al.(2015{\natexlab{b}})Ma, Smith, Jaggi, Jordan, Richt{\'a}rik,
  and Tak{\'a}{\v{c}}]{ma2015adding}
Ma, Chenxin, Smith, Virginia, Jaggi, Martin, Jordan, Michael~I, Richt{\'a}rik,
  Peter, and Tak{\'a}{\v{c}}, Martin.
\newblock Adding vs. averaging in distributed primal-dual optimization.
\newblock In \emph{ICML 2015 - Proceedings of the 32th International Conference
  on Machine Learning}, volume~37, pp.\  1973--1982. JMLR, 2015{\natexlab{b}}.

\bibitem[Marecek et~al.(2014)Marecek, Richt{\'a}rik, and
  Tak{\'a}c]{marecek2014distributed}
Marecek, Jakub, Richt{\'a}rik, Peter, and Tak{\'a}c, Martin.
\newblock Distributed block coordinate descent for minimizing partially
  separable functions.
\newblock \emph{Numerical Analysis and Optimization 2014, Springer Proceedings
  in Mathematics and Statistics}, 2014.

\bibitem[Nitanda(2014)]{nitanda2014stochastic}
Nitanda, Atsushi.
\newblock Stochastic proximal gradient descent with acceleration techniques.
\newblock In \emph{Advances in Neural Information Processing Systems}, pp.\
  1574--1582, 2014.

\bibitem[Niu et~al.(2011)Niu, Recht, R{\'e}, and Wright]{Niu:2011wo}
Niu, Feng, Recht, Benjamin, R{\'e}, Christopher, and Wright, Stephen~J.
\newblock {Hogwild!: A Lock-Free Approach to Parallelizing Stochastic Gradient
  Descent}.
\newblock In \emph{NIPS}, 2011.

\bibitem[Press et~al.(2007)Press, Teukolsky, Vetterling, and
  Flannery]{press2007numerical}
Press, William~H, Teukolsky, Saul~A, Vetterling, William~T, and Flannery,
  Brian~P.
\newblock Numerical recipes: the art of scientific computing.
\newblock 2007.

\bibitem[Qu et~al.(2015)Qu, Richt{\'a}rik, and Zhang]{qu2015quartz}
Qu, Zheng, Richt{\'a}rik, Peter, and Zhang, Tong.
\newblock Quartz: Randomized dual coordinate ascent with arbitrary sampling.
\newblock In \emph{Advances in Neural Information Processing Systems}, pp.\
  865--873, 2015.

\bibitem[Recht et~al.(2011)Recht, Re, Wright, and Niu]{recht2011hogwild}
Recht, Benjamin, Re, Christopher, Wright, Stephen, and Niu, Feng.
\newblock Hogwild: A lock-free approach to parallelizing stochastic gradient
  descent.
\newblock In \emph{Advances in Neural Information Processing Systems}, pp.\
  693--701, 2011.

\bibitem[Richt{\'a}rik \& Tak{\'a}{\v{c}}(2013)Richt{\'a}rik and
  Tak{\'a}{\v{c}}]{richtarik2013distributed}
Richt{\'a}rik, Peter and Tak{\'a}{\v{c}}, Martin.
\newblock Distributed coordinate descent method for learning with big data.
\newblock \emph{arXiv preprint arXiv:1310.2059}, 2013.

\bibitem[Rodgers(1985)]{rodgers1985improvements}
Rodgers, David~P.
\newblock Improvements in multiprocessor system design.
\newblock In \emph{ACM SIGARCH Computer Architecture News}, volume~13, pp.\
  225--231. IEEE Computer Society Press, 1985.

\bibitem[Roux et~al.(2012)Roux, Schmidt, and Bach]{roux2012stochastic}
Roux, Nicolas~L, Schmidt, Mark, and Bach, Francis~R.
\newblock A stochastic gradient method with an exponential convergence \_rate
  for finite training sets.
\newblock In \emph{Advances in Neural Information Processing Systems}, pp.\
  2663--2671, 2012.

\bibitem[Schmidt et~al.(2013)Schmidt, Roux, and Bach]{schmidt2013minimizing}
Schmidt, Mark, Roux, Nicolas~Le, and Bach, Francis.
\newblock Minimizing finite sums with the stochastic average gradient.
\newblock \emph{arXiv:1309.2388}, 2013.

\bibitem[Shalev-Shwartz et~al.(2011)Shalev-Shwartz, Singer, Srebro, and
  Cotter]{shalev2011pegasos}
Shalev-Shwartz, Shai, Singer, Yoram, Srebro, Nathan, and Cotter, Andrew.
\newblock Pegasos: Primal estimated sub-gradient solver for svm.
\newblock \emph{Mathematical programming}, 127\penalty0 (1):\penalty0 3--30,
  2011.

\bibitem[Shamir \& Srebro(2014)Shamir and Srebro]{shamir2014distributed}
Shamir, Ohad and Srebro, Nathan.
\newblock Distributed stochastic optimization and learning.
\newblock In \emph{Communication, Control, and Computing (Allerton), 2014 52nd
  Annual Allerton Conference on}, pp.\  850--857. IEEE, 2014.

\bibitem[Shamir et~al.(2013)Shamir, Srebro, and Zhang]{shamir2013communication}
Shamir, Ohad, Srebro, Nathan, and Zhang, Tong.
\newblock Communication efficient distributed optimization using an approximate
  newton-type method.
\newblock \emph{arXiv preprint arXiv:1312.7853}, 2013.

\bibitem[Tak{\'a}{\v{c}} et~al.(2013)Tak{\'a}{\v{c}}, Bijral, Richt{\'a}rik,
  and Srebro]{takavc2013mini}
Tak{\'a}{\v{c}}, Martin, Bijral, Avleen, Richt{\'a}rik, Peter, and Srebro,
  Nathan.
\newblock Mini-batch primal and dual methods for {SVMs}.
\newblock \emph{ICML}, 2013.

\bibitem[Tak{\'a}{\v{c}} et~al.(2015)Tak{\'a}{\v{c}}, Richt{\'a}rik, and
  Srebro]{takavc2015distributed}
Tak{\'a}{\v{c}}, Martin, Richt{\'a}rik, Peter, and Srebro, Nathan.
\newblock Distributed mini-batch {SDCA}.
\newblock \emph{arXiv preprint arXiv:1507.08322}, 2015.

\bibitem[Yang(2013)]{yang2013trading}
Yang, Tianbao.
\newblock Trading computation for communication: Distributed stochastic dual
  coordinate ascent.
\newblock In \emph{Advances in Neural Information Processing Systems}, pp.\
  629--637, 2013.

\bibitem[Yang et~al.(2013)Yang, Zhu, Jin, and Lin]{yang2013analysis}
Yang, Tianbao, Zhu, Shenghuo, Jin, Rong, and Lin, Yuanqing.
\newblock Analysis of distributed stochastic dual coordinate ascent.
\newblock \emph{arXiv preprint arXiv:1312.1031}, 2013.

\bibitem[Zhang \& Xiao(2015)Zhang and Xiao]{zhang2015communication}
Zhang, Yuchen and Xiao, Lin.
\newblock Communication-efficient distributed optimization of self-concordant
  empirical loss.
\newblock \emph{arXiv preprint arXiv:1501.00263}, 2015.

\end{thebibliography}
\bibliographystyle{icml2015}

\end{document}